\documentclass{article}

\usepackage[accepted]{icml2023}
\usepackage{microtype}
\usepackage{graphicx}
\usepackage{subcaption}
\usepackage{booktabs} % for professional tables

\usepackage{makecell}
\usepackage{hyperref}
\usepackage{amsmath}
\usepackage{amssymb}
\usepackage{mathtools}
\usepackage{microtype}      % microtypography
\usepackage{enumitem}
\usepackage{amsmath,amsthm,amssymb,amsfonts}
\usepackage[capitalize,noabbrev]{cleveref}
\usepackage{xspace}
\usepackage[textsize=tiny]{todonotes}
\usepackage{tikz}
\usepackage{float}
\usepackage{bbm}
\usepackage{enumitem}

\newcommand{\tsn}[1]{{\left\vert\kern-0.25ex\left\vert\kern-0.25ex\left\vert #1 
    \right\vert\kern-0.25ex\right\vert\kern-0.25ex\right\vert}}
\usepackage{xcolor}
\definecolor{darkred}{RGB}{150,0,0}
\definecolor{darkgreen}{RGB}{0,150,0}
\definecolor{darkblue}{RGB}{0,0,200}
\newcommand{\beq}{\begin{equation}}

\newcommand{\eeq}{\end{equation}}

\newcommand{\s}{\vct{s}}
\renewcommand{\d}{\mathrm{d}}

\renewcommand{\P}{\operatorname{\mathbb{P}}}
\def \endprf{\hfill {\vrule height6pt width6pt depth0pt}\medskip}
\usepackage{amsmath,amsfonts,bm}

\def\eqref#1{equation~\ref{#1}}
\def\1{\bm{1}}

\DeclareMathAlphabet{\mathsfit}{\encodingdefault}{\sfdefault}{m}{sl}
\SetMathAlphabet{\mathsfit}{bold}{\encodingdefault}{\sfdefault}{bx}{n}

\newcommand{\E}{\mathbb{E}}

\usetikzlibrary{shapes.arrows, positioning}

\theoremstyle{plain}

\theoremstyle{definition}

\theoremstyle{remark}

\newcommand{\pgen}{{\textsc{PureGen}}\xspace}

\newcommand{\epic}{{\textsc{EPIc}}\xspace}
\newcommand{\pebm}{{\textsc{PureEBM}}\xspace}
\newcommand{\friends}{{\textsc{FrieNDs}}\xspace}

\def\s{{\frac{1}{n} \sum_{i=1}^{n}}}
\def\d{{\nabla_\theta}}

\def\P{{p_{\rm data}}}

\icmltitlerunning{\pebm: Universal Poison Purification via Mid-Run Dynamics of Energy-Based Models}

\begin{document}

\twocolumn[
\icmltitle{\pebm: Universal Poison Purification via Mid-Run Dynamics of Energy-Based Models}

\icmlsetsymbol{equal}{*}

\begin{icmlauthorlist}
\icmlauthor{Omead Pooladzandi}{equal,xxx}
\icmlauthor{Jeffrey Jiang}{equal,yyy}
\icmlauthor{Sunay Bhat}{equal,yyy}
\icmlauthor{Gregory Pottie}{yyy}
\end{icmlauthorlist}

\icmlaffiliation{xxx}{Department of Electrical Engineering, California Institute of Technology, Pasadena}
\icmlaffiliation{yyy}{Department of Electrical and Computer Engineering, University of California, Los Angeles}

\icmlcorrespondingauthor{Sunay Bhat}{sunaybhat1@ucla.edu}
\icmlcorrespondingauthor{Omead Pooladzandi}{omead@caltech.edu}

\icmlkeywords{Energy Based Models, Poison Defense}

\vskip 0.3in
]



\printAffiliationsAndNotice{\icmlEqualContribution} % otherwise use the standard text.

\begin{abstract}
%% Condensed
Data poisoning attacks pose a significant threat to the integrity of machine learning models by leading to misclassification of target distribution data by injecting adversarial examples during training. Existing state-of-the-art (SoTA) defense methods suffer from limitations, such as significantly reduced generalization performance and significant overhead during training, making them impractical or limited for real-world applications. In response to this challenge, we introduce a universal data purification method that defends naturally trained classifiers from malicious white-, gray-, and black-box image poisons by applying a universal stochastic preprocessing step $\Psi_{T}(x)$, realized by iterative Langevin sampling of a convergent Energy Based Model (EBM) initialized with an image $x.$ Mid-run dynamics of $\Psi_{T}(x)$ purify poison information with minimal impact on features important to the generalization of a classifier network. We show that EBMs remain universal purifiers, even in the presence of poisoned EBM training data, and achieve SoTA defense on leading triggered and triggerless poisons. This work is a subset of a larger framework introduced in \pgen with a more detailed focus on EBM purification and poison defense. We make our code available on GitHub.\footnote{\url{https://github.com/SunayBhat1/PureGen_PoisonDefense}}
\end{abstract}

\section{Introduction}
\begin{figure*}[ht]
    \begin{subfigure}[b]{\textwidth}
        \centering
        \scalebox{0.65}{\input{EBM/process_diagram}}
        \label{fig:process_diagram}
        % \caption{PureEBM Process}
    \end{subfigure}
    % \vspace{-3mm}
    % \centering
    % \vspace{-5pt}
    % \begin{subfigure}[b]{0.9\linewidth}
    % \centering
    \begin{tikzpicture}
        \node[anchor=south west,inner sep=0] (image) at (0,0) {
            \includegraphics[width=0.5\textwidth]{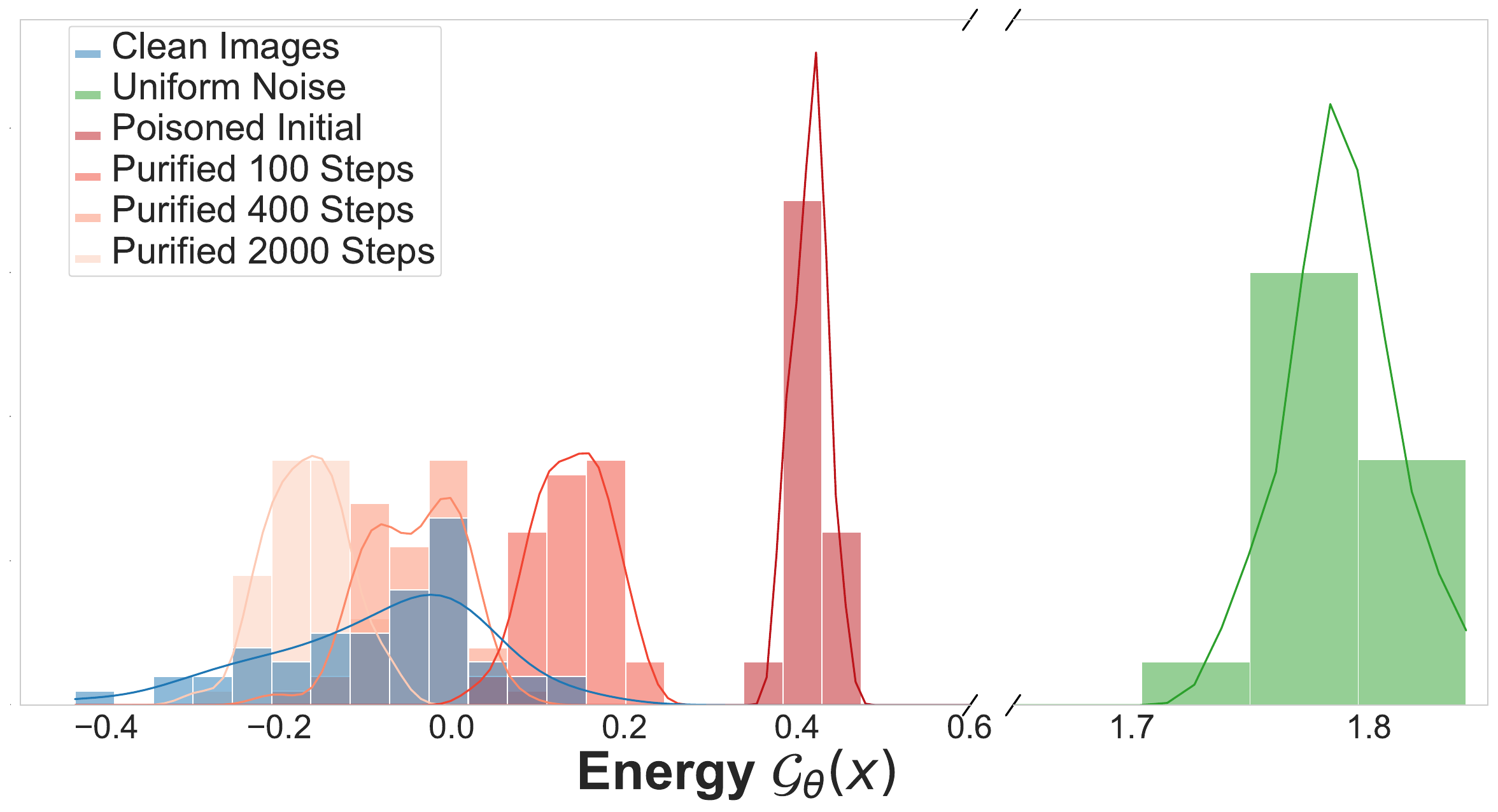}
        };
        % Draw a fat arrow
        \draw [-latex, line width=2mm, teal!45] (4.5,3.5) -- (1.5,2);
        % Place a text box
        \node [draw, fill=white, text width=1.8cm, align=center, above right=3.9cm and 2.8cm of image.south west] (textbox) {
            \pebm{}
        };
    \vspace{-15mm}
    \end{tikzpicture}
    % \end{subfigure}
    % \vspace{-5pt}
    % Bottom row: one wide image
    % \vspace{-10mm}
    \begin{subfigure}[b]{\textwidth}
        % \centering
        \includegraphics[width=0.5\textwidth]{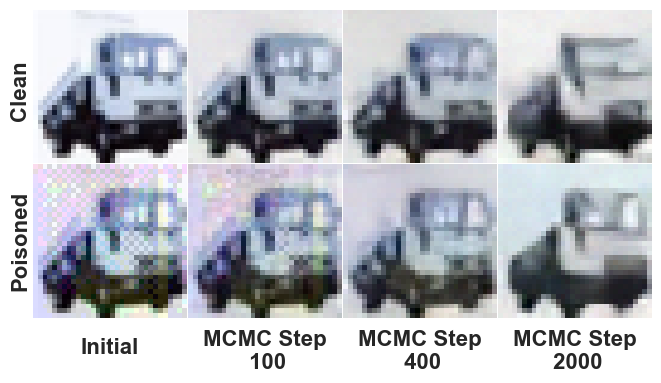}
        \label{fig:purified_images_intro}
        % \caption{PureEBM Process}
    \end{subfigure}
    \vspace{-15pt}
    \caption{\textbf{Top} The full \pebm pipeline is shown where we apply our method as a preprocessing step with no further downstream changes to the classifier training or inference. \textit{Poisoned images are moderately exaggerated to show visually.} \textbf{Bottom Left} Energy distributions of clean, poisoned, and purified images. Our method pushes poisoned images via purification into the natural image energy manifold. \textbf{Bottom Right} The removal of poisons and similarity of clean and poisoned images with more MCMC steps. The purified dataset results in SoTA defense and high classifier accuracy.}
    \label{fig:intro_figure}
    % \vspace{-2mm}
\end{figure*}

Large datasets empower modern, over-parameterized deep learning models. An adversary can easily insert a small number of powerful, but imperceptible, poisoned images into these datasets, often scraped from the open Internet, and manipulate a Neural Network's (NN) behavior at test time with a high success rate. These poisons can be constructed with or without information on NN architecture or training dynamics. With the increasing capabilities and utilization of larg deep learning models, there is growing research in securing model training against such adversarial poison attacks with minimal impact on natural accuracy.

Numerous methods of poisoning deep learning systems have been proposed in recent years. These disruptive techniques typically fall into two distinct categories: backdoor, triggered data poisoning, or triggerless poisoning attacks. Triggered attacks  conceal an imperceptible trigger pattern in the samples of the training data leading to the misclassification of test-time samples that contain the hidden trigger \cite{gu2017badnets,turner2018clean,souri2021sleeper,zeng2022narcissus}. In contrast, triggerless poisoning attacks involve introducing slight, bounded perturbations to individual images that align them with target images of another class within the feature or gradient space resulting in the misclassification of specific instances without necessitating further modification during inference \cite{Shafahi2018poisonfrogs,zhu2019transferable,huang2020metapoison,geiping2021witches,aghakhani2021bullseye}. In both scenarios, poisoned examples often appear benign and correctly labeled, making them challenging to detect by observers or algorithms.

Current defense strategies against data poisoning exhibit significant limitations. While some methods rely on anomaly detection through techniques such as nearest neighbor analysis, training loss minimization, singular-value decomposition, feature activation or gradient clustering \cite{cretu2008casting, Steinhardt17certified,tran2018spectral,chen2019detecting,peri2020deep,yang2022poisons,pooladzandi2022adaptive,omead_thesis}, others resort to robust training strategies including data augmentation, randomized smoothing, ensembling, adversarial training and maximal noise augmentation \cite{weber2020rab,levine2020deep,abadi2016deep,ma2019data,li2021anti,tao2021better,liu2023friendly}. However, these approaches either undermine the model's generalization performance \citep{geiping2021doesn,yang2022poisons}, offer protection only against specific attack types \citep{geiping2021doesn,peri2020deep,tran2018spectral}, or prove computationally prohibitive for standard deep learning workflows \citep{abadi2016deep,chen2019detecting,madry2018towards,yang2022poisons,geiping2021doesn,peri2020deep,liu2023friendly}. There remains a critical need for more effective and practical defense mechanisms in the realm of deep learning security.

In this work, we propose a simple but powerful Energy-Based model defense \pebm, against poisoning attacks. We make the key observation that the energy of poisoned images is significantly higher than that of baseline images for an EBM trained on a natural dataset of images (even when poisoned samples are present). Using iterative sampling techniques such as Markov Chain Monte Carlo (MCMC) that utilize noisy gradient information from the EBM, we can purify samples of any poison perturbations iteratively. This universal stochastic preprocessing step $\Psi_{T}(x)$ moves poisoned samples into the lower energy, natural data manifold with minimal loss in natural accuracy. The \pebm pipeline, energy distributions, and the MCMC purification process on a sample image can be seen in Figure \ref{fig:intro_figure}. This work finds that \pebm significantly outperforms state-of-the-art defense methods in all tested poison scenarios. Our key contributions in this work are:

\vspace{-2mm}

\begin{itemize} %[itemsep=0.2pt, topsep=0pt]
    \itemsep0em
    \item A state-of-the-art stochastic preprocessing defense $\Psi_{T}(x)$ against adversarial poisons, using Energy-Based models and MCMC sampling
    \item Experimental results showing the broad application of $\Psi_{T}(x)$ with minimal tuning and no prior knowledge needed of the poison type and classification model 
    \item Results showing SoTA performance is maintained when the EBM training data includes poisoned samples and/or natural images from a similar out-of-distribution dataset
\end{itemize}

\vspace{-5mm}
\section{Related Work}
% In this section we will cover basics on both poisoning as well as SoTA defenses. 

% Typically a stochastic gradient descent process is utilized over some objective to maximize either the inference time impact of a trigger, or modify a target images classification by aligning poisoned images in the gradient or some feature space. 

\subsection{Targeted Data Poisoning Attack} 
% for some patch $x_{patch}$ for poisoned class $C_p$, we change some small subset $S \subseteq C_p$ such that $|S| = \lambda$ and replace the values of $S$ with the values of $P = \{ x + x_{patch} : x \in S\}$ in training. 
Poisoning of a dataset occurs when an attacker injects small adversarial perturbations $\delta$ (where  $\| \delta \|_{\infty}$ $\leq$ $\xi$ and typically $\xi=8/255$) into a small fraction, $\alpha$, of training images. These train-time attacks introduce \textit{local sharp regions} with a considerably higher \textit{training loss} \cite{liu2023friendly}. A successful attack occurs when SGD optimizes the cross-entropy training objective on these poisoned images, maximizing either the inference time impact of a trigger, or modifying a target image classification by aligning poisoned images in the gradient or some feature space. The process of learning these adversarial perturbations creates backdoors in an NN.

In the realm of deep network poison security, we encounter two primary categories of attacks: triggered and triggerless attacks. Triggered attacks, often referred to as backdoor attacks, involve contaminating a limited number of training data samples with a specific trigger (often a patch) $\rho$ (similarly constrained $\| \rho \|_{\infty}$ $\leq$ $\xi$) that corresponds to a target label, $y^{\text{adv}}$. After training, a successful backdoor attack misclassifies when the perturbation $\rho$ is added:
\begin{equation}
    F(x) = \begin{cases}
    y & x \in \{x : (x, y) \in \mathcal{D}_{test}\} \\
    y^{\text{adv}} & x \in \{ x + \rho : (x, y) \in \mathcal{D}_{test}, y \ne y^\text{adv} \}
\end{cases}
\end{equation}
Early backdoor attacks were characterized by their use of non-clean labels~\cite{chen2017targeted, gu2017badnets, liu2017trojaning, souri2021sleeper}, but more recent iterations of backdoor attacks have evolved to produce poisoned examples that lack a visible trigger~\cite{turner2018clean, Saha2019htbd,zeng2022narcissus}.

On the other hand, triggerless poisoning attacks involve the addition of subtle adversarial perturbations to base images, aiming to align their feature representations or gradients with those of target images of another class, causing target misclassification~\cite{Shafahi2018poisonfrogs, zhu2019transferable, huang2020metapoison, geiping2021witches, aghakhani2021bullseye}. These poisoned images are virtually undetectable by external observers. Remarkably, they do not necessitate any alterations to the target images or labels during the inference stage. For a poison targeting a group of target images $\Pi = \{ (x^\pi , y^\pi) \}$ to be misclassified as $y^{\text{adv}}$, an ideal triggerless attack would produce a resultant function:
\begin{equation}
    F(x) = \begin{cases}
    y & x \in \{ x : (x, y) \in \mathcal{D}_{test} \setminus \Pi\} \\
    y^{\text{adv}} & x \in \{ x: (x, y) \in \Pi \} \\
\end{cases}
\end{equation}

The current leading poisoning attacks that we assess our defense against are:
\vspace{-2mm}
\begin{itemize} 
\itemsep0em 
  \item \textbf{Bullseye Polytope (BP):} BP crafts poisoned samples that position the target near the center of their convex hull in a feature space \cite{aghakhani2021bullseye}.

  \item \textbf{Gradient Matching (GM):} GM generates poisoned data by approximating a bi-level objective by aligning the gradients of clean-label poisoned data with those of the adversarially labeled target~\cite{geiping2021witches}. This attack has shown effectiveness against data augmentation and differential privacy.

  \item \textbf{Narcissus (NS):} NS is a clean-label backdoor attack that operates with minimal knowledge of the training set, instead using a larger natural dataset, evading state-of-the-art defenses by synthesizing persistent trigger features for a given target class.~\cite{zeng2022narcissus}.
\end{itemize}

\subsection{Defense Strategies} 

Poison defense categories broadly take two primary approaches: filtering and robust training techniques. Filtering methods identify outliers in the feature space through methods such as thresholding \cite{Steinhardt17certified}, nearest neighbor analysis \cite{peri2020deep}, activation space inspection \cite{chen2019detecting}, or by examining the covariance matrix of features \cite{tran2018spectral}. These defenses often assume that only a small subset of the data is poisoned, making them vulnerable to attacks involving a higher concentration of poisoned points. Furthermore, these methods substantially increase training time, as they require training with poisoned data, followed by computationally expensive filtering and model retraining \cite{chen2019detecting,peri2020deep,Steinhardt17certified,tran2018spectral}. 

On the other hand, robust training methods involve techniques like randomized smoothing \cite{weber2020rab}, extensive data augmentation \cite{borgnia2021strong}, model ensembling \cite{levine2020deep}, gradient magnitude and direction constraints \cite{hong2020effectiveness}, poison detection through gradient ascent \cite{li2021anti}, and adversarial training \cite{geiping2021doesn,madry2018towards,tao2021better}. Additionally, differentially private (DP) training methods have been explored as a defense against data poisoning \cite{abadi2016deep,jayaraman2019evaluating}. Robust training techniques often require a trade-off between generalization and poison success rate \cite{abadi2016deep,hong2020effectiveness,li2021anti,madry2018towards,tao2021better,liu2023friendly} and can be computationally intensive \cite{geiping2021doesn,madry2018towards}. Some methods use optimized noise constructed via Generative Adversarial Networks (GANs) or Stochastic Gradient Descent methods to make noise that defends against attacks \cite{ madaan2021learning,liu2023friendly}.

Recently \citet{yang2022poisons} proposed \epic, a coreset selection method that rejects poisoned images that are isolated in the gradient space throughout training, and \citep{liu2023friendly} proposed \friends, a per-image pre-processing transformation that solves a min-max problem to stochastically add $l_{\infty}$ norm $\zeta$-bound `friendly noise' (typically 16/255) to combat adversarial perturbations. These two methods are the SoTA and will serve as a benchmark for our \pebm method in the experimental results. 
 
% We implement EPIC and FrieNDs \cite{yang2022poisons,liu2023friendly} as the leading data filtering and data augmentation techniques respectively, as baseline defense methods to compare our approach against in experimental results. 
% Epic relies specifically on a CRAIG like clusterting approach to remove that iteratively removes examples with isolated (outlier) gradients every N epochs. Friendly introduces random noise as a data augmentation, and a fixed perturbation obtained through an optimization process designed to maximally perturb the image in some $\epsilon_F$ bound (typically 16) while minimally impacting the model's class preidction probability distibrution some N epochs into training. 

When compared to augmentation-based and adversarial training methods, our approach stands out for its simplicity, speed, and ability to maintain strong generalization performance. We show that adding gradient noise in the form of iterative Langevin updates can purify poisons and achieve superior generalization performance compared to SoTA defense methods \epic and \friends. The Langevin noise in our method proves highly effective in removing the adversarial signals while metastable behaviors preserve features of the original image, due to the dynamics of mid-run chains from our EBM defense method.

\section{\pebm: Purifying Langevin Defense against Poisoning Attacks }
% \vspace{-5mm}
Given a clean training set $\mathcal{X}_{clean} \subset \mathbb{R}^D $ consisting of i.i.d. sample images $x_i \sim p_{clean}$ for $i = 1,\dots,n$. 
Targeted data poisoning attacks modify $\alpha n$ training points, by adding optimized perturbations $\delta$ constrained by $\mathcal{C} = \{ \delta \in \mathbb{R}^D: \| \delta \|_\infty \le \xi \}$. Poisons crafted by such attacks look innocuous to human observers and are seemingly labeled correctly. Hence, they are called clean-label attacks. These images define a new distribution $x_i + \delta_i \sim p_{poison}$, so that our training set comes from the mixture of probability distributions: 
\begin{equation}
    % \vspace{-5mm}
    % \vspace{1mm}
    p_{data} = (1-\alpha)p_{clean} + \alpha p_{poison}
    \label{eq:mixture}
    % \vspace{-5mm}
\end{equation}
The goal of adding these poisons is to change the prediction of a set of target examples $\Pi = \{(x^{\pi},y^{\pi})\} \subset \mathcal{D}_{test}$ or triggered examples $\{(x + \rho,y): (x, y) \in \mathcal{D}_{test} \}$ to an adversarial label $y^{\text{adv}}$. 

Targeted clean-label data poisoning attacks can be formulated as the following bi-level optimization problem: 
\begin{align}\label{eq:poisoning}
    % \vspace{-5mm}
    \underset{\substack{ {\delta_i \in\mathcal{C}_\delta, \rho \in \mathcal{C}_{\rho}} \\ {\sum_{i=0}^n\mathbbm{1}_{\delta_i \ne \mathbf{0}} \le \alpha n}} }{\arg\!\min} & \sum_{(x^{\pi},y^{\pi}) \in \Pi}\mathcal{L}\left(F(x^{\pi} + \rho; \phi(\delta)), y^{\text{adv}} \right) \nonumber \\
     s.t.\quad  \phi (\delta)\! &=\! \underset{\phi}{\arg\!\min} \sum_{(x,y) \in \mathcal{D}} \mathcal{L}\left(F(x\!+\!\delta_i; \phi), y\right) 
     % \vspace{-5mm}
\end{align}
For a triggerless poison, we solve for the ideal perturbations $\delta_i$ to minimize the adversarial loss on the target images, where $\mathcal{C}_\delta = \mathcal{C}$, $\mathcal{C}_\rho = \{ \mathbf{0} \in \mathbb{R}^D \}$, and $\mathcal{D} = \mathcal{D}_{train} $. 
To address the above optimization problem, powerful poisoning attacks such as Meta Poison (MP) \citep{huang2020metapoison}, Gradient Matching (GM) \cite{geiping2021witches}, and Bullseye Polytope (BP) \cite{aghakhani2021bullseye} craft the poisons to mimic the gradient of the adversarially labeled target, i.e.,
% \begin{equation}\label{eq:matching}
%     \nabla\mathcal{L}(x^{\pi},y^{\text{adv}},\phi)\approx\frac{1}{n-m}\sum_{i=m}^{n} \nabla\mathcal{L}(F(x_i+\delta_i,\phi), y_i),
% \end{equation} 
\begin{align}\label{eq:matching}
    \!\!\nabla\mathcal{L}\left(F_\phi\left(x^{\pi}\right),y^{\text{adv}}\right)\propto \!\sum_{i:\delta_i \ne \mathbf{0}} \!\nabla\mathcal{L}\left(F_\phi(x_i+\delta_i), y_i\right)
    % \vspace{-5mm}
\end{align}
Minimizing the training loss on RHS of Equation \ref{eq:matching}
also minimizes the adversarial loss objective of Equation \ref{eq:poisoning}. 
% \je{Small inconsistency: Equation 5 assumes single target $x^\pi$, while equation 4 still assumes multiple $x^\pi$s. Do we want to mention 1 target or change eq 5 to average over all $x^\pi$s?}
% \om{the more general the better i think}

For the triggered poison, Narcissus (NS), we find the most representative patch $\rho$ for class $\pi$ given $\mathcal{C}$, defining Equation \ref{eq:poisoning} with $\mathcal{C}_\delta = \{ \mathbf{0} \in \mathbb{R}^D \} $, $\mathcal{C}_\rho\!= \mathcal{C}$, $\Pi = \mathcal{D}_{train}^\pi, y^{\text{adv}} = y^\pi $, and $\mathcal{D} = \mathcal{D}_{POOD} \cup \mathcal{D}_{train}^\pi $. In particular, this patch uses a public out-of-distribution dataset $\mathcal{D}_{POOD}$ and only the targeted class $\mathcal{D}_{train}^\pi$. As finding this patch comes from another natural dataset and does not depend on other train classes, NS has been more flexible to model architecture, dataset, and training regime \cite{zeng2022narcissus}.

\subsection{Energy-Based Model}\label{sec:ebm_1}
An Energy-Based Model (EBM) is formulated as a Gibbs-Boltzmann density, as introduced in \citep{xie2016theory}. This model can be mathematically represented as:
\begin{equation}
    p_{\theta}(x) = \frac{1}{Z(\theta)}\exp(-\mathcal{G}_{\theta}(x))q(x), \label{eq:ebm}
\end{equation}
where $x \in \mathcal{X} \subset \mathbb{R}^D$ denotes an image signal, and $q(x)$ is a reference 
measure, often a uniform or standard normal distribution. Here, $\mathcal{G}_{\theta}$ signifies the 
energy potential, parameterized by a ConvNet with parameters $\theta$. The normalizing constant, 
or the partition function, $Z(\theta) = \int \exp \{-\mathcal{G}_{\theta}(x) \} q(x) dx = \E_q[\exp(-\mathcal{G}_{\theta}(x))]$, while essential, is generally analytically intractable. In practice, $Z(\theta)$ is not computed explicitly, as $\mathcal{G}_{\theta}(x)$ sufficiently informs the Markov Chain Monte Carlo (MCMC) sampling process.

As which $\alpha$ of the images are poisoned is unknown, we treat them all the same for a universal defense. Considering i.i.d. samples $x_i\sim \P$ for $i = 1, \dots ,n$, with $n$ sufficiently large, the sample average over ${x_i}$ converges to the expectation under $\P$ and one can learn a parameter $\theta^*$ such that $p_{\theta^*}(x) \approx \P(x) $. For notational simplicity, we equate the sample average with the expectation. 

The objective is to minimize the expected negative log-likelihood, formulated as:
\begin{equation}
\mathcal{L}(\theta)= \s \log p_{\theta}(x_i) \doteq \E_{\P}[ \log p_\theta(x)].
\label{eq:loglikelihood}
\end{equation}
The derivative of this log-likelihood, crucial for parameter updates, is given by:
\begin{align}\nonumber
\nabla \mathcal{L}(\theta) &= \E_{\P} \left[ \d \mathcal{G}_{\theta}(x)\right] - \E_{p_{\theta}} \left[ \d \mathcal{G}_{\theta}(x) \right] \\
&\doteq \s \d \mathcal{G}_{\theta}(x_i^+) - \frac{1}{k}\sum_{i=1}^{k} \d \mathcal{G}_{\theta}(x_i^{-}),
\label{eq:theta_update}
\end{align}
where $x_i^{+}$ are called \textit{positive} samples as their probability is increased and where  $k$ samples $x_i^{-} \sim p_{\theta}(x)$ are synthesized examples (obtained via MCMC) from the current model, representing the \textit{negative} samples as probability is deceased. 
% \je{I think prob increase/decrease feels slightly confusing to reader}

 In each iteration $t$, with current parameters denoted as ${\theta}_t$, we generate $k$ synthesized
 examples $x_i^{-} \sim p_{\theta_t}(x)$. The parameters are then updated as $\theta_{t+1} = \theta_t + \eta_t \nabla \mathcal{L}(\theta_t)$, where $\eta_t$ is the learning rate. 

 In this work, to obtain the negative samples $x_i^{-}$ from the current distribution $p_{\theta}(x)$ we utilize the iterative application of the Langevin update as the MCMC method:
\begin{align}
x_{\tau+1} = x_{\tau} - \Delta \tau \nabla_{x_{\tau}} \mathcal{G}_{\theta}(x_{\tau}) + \sqrt{2\Delta \tau} \epsilon_{\tau}  ,
\label{eqn:langevin}
\end{align}
where $\epsilon_k \sim \text{N}(0, I_D )$, $\tau$ indexes the time step of the Langevin dynamics, and $\Delta \tau$ is the discretization of time \cite{xie2016theory}. $\nabla_{x} \mathcal{G}_{\theta}(x) = \partial \mathcal{G}_{\theta}(x)/ \partial x$ can be obtained by back-propagation. If the gradient term dominates the diffusion noise term, the Langevin dynamics behave like gradient descent. We implement EBM training following \cite{nijkamp2019anatomy}, see App \ref{app:ebm_train} for details.

\begin{algorithm}
    \caption{Data Preprocessing with \pebm:  $\Psi_{T}(x)$ }

\begin{algorithmic}
\REQUIRE  Trained ConvNet potential $\mathcal{G}_{\theta}(x)$, training images $x\in X$, Langevin steps $T$, Time discretization $\Delta \tau$
    \FOR{$\tau$ in $1 \ldots T$}
        \STATE Langevin Step: draw $\epsilon_{\tau} \sim \text{N}(0, I_D )$ 
            \[ x_{\tau+1} = x_{\tau} - \Delta \tau \nabla_{x_{\tau}} \mathcal{G}_{\theta}(x_{\tau}) + \sqrt{2\Delta \tau} \epsilon_{\tau}   \]
    \ENDFOR
\STATE {\textbf{Return: } Purified set $\Tilde{X}$ from final Langevin updates}
% \RETURN {}
\end{algorithmic}

\label{alg_1}
\end{algorithm}

In practice, we find that learning the mixture of distributions $p_{data} = (1-\alpha)p_{clean} + \alpha p_{poison}$ yields an EBM with a purifying ability similar to that of training on $p_{clean}$, suggesting our unsupervised MLE method is unsurprisingly not affected by targeted poisons. 

\subsection{Classification with Stochastic Transformation}\label{subsec:stochtrans}
Let $\Psi_{T}:\mathbb{R}^D \rightarrow \mathbb{R}^D$ be a stochastic pre-processing transformation. In this work, $\Psi_{T}(x)$, the random variable of a fixed image $x$, is realized via $T$ steps of the Langevin update \eqref{eqn:langevin}. One can compose a stochastic transformation $\Psi_{T}(x)$ with a randomly initialized deterministic classifier $f_{\phi_0}(x)\in \mathbb{R}^J$  (for us, a naturally trained classifier) to define a new deterministic classifier $F_{\phi}(x)\in\mathbb{R}^J$ as 
    $F_{\phi_0}(x) = E_{\Psi_{T}(x)}[f_{\phi_0}(\Psi_{T}(x))],$
which is then trained with cross-entropy loss via SGD to realize $F_{\phi}(x)$. 
% \je{Is $J = N_{class}$ here? }
As it is infeasible to evaluate the above expectation of the stochastic transformations $\Psi_{T}(x)$ as well as training many randomly initialized classifiers we take $f_{\phi}(\Psi_{T}(x))$ as the point estimate of the classifier $F_{\phi}(x)$. In our case this instantaneous approximation of $F_{\phi}(x)$ is valid because $\Psi_{T}(x)$ has a low variance for convergent mid-run MCMC.

% As such, one can simply pre-process the full training dataset once using MCMC and train a natural classifier on the transformed dataset with no modification to the vanilla training process.  
% \om{consider putting short alg here, just SGD with transformation being $\Psi_{T}$ at the start.}

\subsection{Why EBM Langevin Dynamics Purify}\label{subsec:properties}

The theoretical basis for eliminating adversarial signals using MCMC sampling is rooted in the established steady-state convergence characteristic of Markov chains. The Langevin update, as specified in Equation (\ref{eqn:langevin}), converges to the distribution $p_{\theta}(x)$ learned from unlabeled data after an infinite number of Langevin steps. The memoryless nature of a steady-state sampler guarantees that after enough steps, all adversarial signals will be removed from an input sample image. Full mixing between the modes of an EBM will undermine the original natural image class features, making classification impossible \cite{hill2021stochastic}. \citet{nijkamp2019anatomy} reveals that without proper tuning, EBM learning heavily gravitates towards \emph{non-convergent ML} where short-run MCMC samples have a realistic appearance and long-run MCMC samples have unrealistic ones. In this work, we use image initialized \emph{convergent learning}. $p_{\theta}(x)$ is described further by Algorithm \ref{alg_1}. 
% which is essential for our defense strategy because mid-run samples must be realistic (while purifying) so that classifiers can maintain high accuracy after the stochastic Langevin transformation. 

The metastable nature of EBM models exhibits characteristics that permit the removal of adversarial signals while maintaining the natural image's class and appearance \cite{hill2021stochastic}. Metastability guarantees that over a short number of steps, the EBM will sample in a local mode, before mixing between modes. Thus, it will sample from the initial class and not bring class features from other classes in its learned distribution. Consider, for instance, an image of a horse that has been subjected to an adversarial $\ell_{\infty}$ perturbation, intended to deceive a classifier into misidentifying it as a dog. The perturbation, constrained by the $\ell_{\infty}$-norm ball, is insufficient to shift the EBM's recognition of the image away from the horse category. Consequently, during the brief sampling process, the EBM actively replaces the adversarially induced `dog' features with characteristics more typical of horses, as per its learned distribution resulting in an output image resembling a horse more closely than a dog. It is important to note, however, that while the output image aligns more closely with the general characteristics of a horse, it does not precisely replicate the specific horse from the original, unperturbed image.

% Let's say we have an adversarial horse image. The natural image was a horse, and an $l_{\infty}$ perturbation was applied to it to trick the classifier into thinking it was a dog. Because of the $l_{\infty}$ norm ball perturbation limit, the EBM is still going to recognize the image as a horse. Under a short number of steps, the EBM will replace the adversarial ``dog'' features with features of horses in its learned distribution, which will make it look more like a horse than a dog, even though it won't represent the exact horse from the original natural image.

Our experiments show that the mid-run trajectories (100-1000 MCMC steps) we use to preprocess the dataset $\mathcal{X}$ capitalize on these metastable properties by effectively purifying poisons while retaining high natural accuracy on $F_{\phi}(x)$ with no training modification needed. A chaos theory-based perspective on EBM dynamics can be found in App. \ref{sec:chaos}.

% Instead, the quasi-equilibrium and meta-stable behaviors of MCMC sampling can explain why the iterative Langevin update can preserve class features while erasing adversarial signals \cite{hill2021stochastic}. Meta-stable behavior refers to when the sampling process (trajectory) tends to remain within a particular mode (or distribution) for a significant duration before transitioning to other modes. In simpler terms, meta-stability implies that the MCMC process gets `stuck' in certain parts of the sample space for intermediate durations. Our method \pebm relies on a balance between the memoryless properties of MCMC sampling that erase noise and the metastable properties of MCMC sampling that preserve the initial state. Successful classification of mid-run MCMC samples occurs when the metastable regions of the EBM $p_{\theta}(x)$ are aligned with the class boundaries learned by the classifier network $F_{\phi}(x)$.

% This is in contrast to the ideal behavior of MCMC, where samples should, over very long time scales, be able to move freely between all modes (intermodal mixing) of the distribution being sampled. 

% Our transformed samples exhibit a strong dependence on the initial state where mixing within a metastable region can greatly reduce the influence of an initial adversarial signal even when full mixing is not occurring. 

% \begin{algorithm}
%     \caption{\pebm: Defining $\Psi_{T}(x)$ Purification }
%     \input{EBM/Algos/purify_data}
% \end{algorithm}

\subsection{Erasing Poison Signals via Mid-Run MCMC}\label{sec:mcmc_dynamics}

% This section discusses an intuitive perspective with empirical results that justify the use of an EBM for purifying poison signals via short to mid-run sampling dynamics. 

The stochastic transform $\Psi_{T}(x)$ is an iterative process, akin to a noisy gradient descent, over the unconditional energy landscape of a learned data distribution. As MCMC is run, the images will move from their initial energy toward $p_{data}$. As shown in Figure \ref{fig:intro_figure}, the energy distributions of poisoned images are much higher, pushing the poisons away from the likely manifold of natural images. By using mid-run dynamics (150-1000 Langevin steps), we transport poisoned images back toward the center of the energy basin. 

\begin{figure}[ht]
    \centering
    \includegraphics[width=\linewidth]{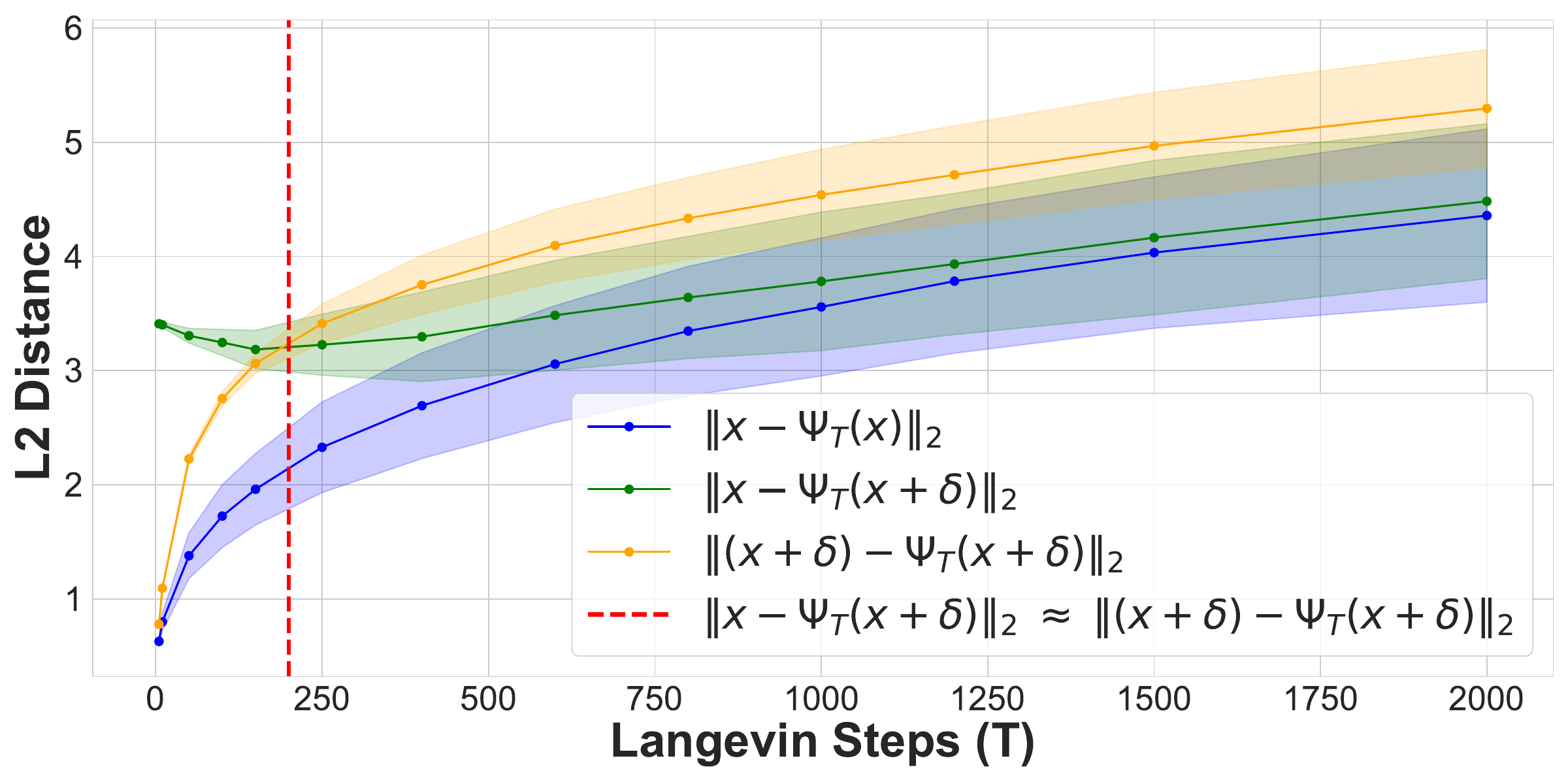}
    % \vspace{-6mm}
    \caption{Plot of $\ell_2$ distances between clean images and clean purified (blue), clean images and poisoned purified (green), and poisoned images and poisoned purified images (orange) at points on the MCMC sampling trajectory. 
    % \je{Pottie says up to here is enough} 
    Purifying poisoned images for less than 250 steps moves a poisoned image closer to its clean image with a minimum around 150, preserving the natural image while removing the adversarial features. 
    % Purified clean and purified poisoned images converge to similar distances away from clean (blue and green lines). The purified poisons move further away from the poisoned starting image (orange line) and intersect with the green line when poisoned samples are closer to the clean image than the poison starting point (dotted red line).
    }
    \vspace{-10pt}
    \label{fig:ebm_L2_dynamics}
\end{figure}

In the from-scratch poison scenarios, 150 Langevin steps can fully purify the majority of the dataset with minimal feature loss to the original image. In Figure \ref{fig:ebm_L2_dynamics} we explore the MCMC trajectory's impacts on $\ell_2$ distance of both purified clean and poisoned images from the initial clean image ($\| x - \Psi_{T}(x) \|_2$ and $\| x - \Psi_{T}(x + \delta) \|_2$), and the purified poisoned image's trajectory away from its poisoned starting point ($\| (x + \delta) - \Psi_{T}(x + \delta) \|_2$). Both poisoned and clean distance trajectories converge to similar distances away from the original clean image ($\lim_{T \to \infty} \| x - \Psi_{T}(x) \|_2 = \lim_{T \to \infty} \| x - \Psi_{T}(x + \delta) \|_2
$), but the steady increase in image distance of the two trajectories offers an empirical perspective of the metastable, mid-run region. The intersection where $\| (x + \delta) - \Psi_{T}(x + \delta) \|_2 > \| x - \Psi_{T}(x + \delta) \|_2$ (indicated by the dotted red line), occurs at $\sim$150-200 Langevin steps and indicates when purification has moved the poisoned image closer to the original clean image than the poisoned version of the image. This region coincides with the expected start of the mid-run dynamics where our properties are most ideal for purification. Additional purification degrades necessary features for classifier training, as already seen previously in the bottom right of Figure \ref{fig:intro_figure}.

 We note that we are not the first to apply EBMs with MCMC sampling for robust classification, but we are, to the best of our knowledge, the first to apply an EBM-based purification method universally as a poison defense and use non-overlapping natural datasets to further extend the generality of EBM purification.

\section{Experiments}

\subsection{Experimental Details}

\begin{table*}[ht!]
    \centering
    \caption{Poison success and natural accuracy in all poisoned training scenarios (ResNet18, CIFAR-10). We report the mean and the standard deviations (as subscripts) of 100 GM experiments, 50 BP experiments, and NS triggers over 10 classes. }
    % First Table
    \scalebox{0.60}{% Please add the following required packages to your document preamble:

\begin{tabular}{@{}rcccccccccc@{}}
% \vspace{-8mm}
\toprule
\multicolumn{11}{c}{\textbf{From Scratch}}                                                                                                                                                                                                                                                                                                                                                                                                                                                                                                                                                                                                                                                                                                               \\ \midrule
\textbf{}                                                            & \multicolumn{5}{c}{\textbf{200 - Epochs}}                                                                                                                                                                                                                                                                                               & \multicolumn{5}{c}{\textbf{80 - Epochs}}                                                                                                                                                                                                                                                                                                \\ \cmidrule(lr){2-6} \cmidrule(lr){7-11} 
\textbf{}                                                            & \multicolumn{2}{c}{\textbf{Gradient Matching-1\%}}                                                                              & \multicolumn{3}{c}{\textbf{Narcissus-1\%}}                                                                                                                                                            & \multicolumn{2}{c}{\textbf{Gradient Matching-1\%}}                                                                              & \multicolumn{3}{c}{\textbf{Narcissus-1\%}}                                                                                                                                                            \\ \cmidrule(lr){2-3} \cmidrule(lr){4-6} \cmidrule(lr){7-8} \cmidrule(lr){9-11} 
\textbf{}                                                            & \begin{tabular}[c]{@{}c@{}}Poison \\ Success (\%) ↓\end{tabular} & \begin{tabular}[c]{@{}c@{}}Avg Natural \\ Accuracy (\%) ↑\end{tabular} & \begin{tabular}[c]{@{}c@{}}Avg Poison \\ Success (\%) ↓\end{tabular} & \begin{tabular}[c]{@{}c@{}}Avg Natural \\ Accuracy (\%) ↑\end{tabular} & \begin{tabular}[c]{@{}c@{}}Max Poison \\ Success (\%) ↓\end{tabular} & \begin{tabular}[c]{@{}c@{}}Poison \\ Success (\%) ↓\end{tabular} & \begin{tabular}[c]{@{}c@{}}Avg Natural \\ Accuracy (\%) ↑\end{tabular} & \begin{tabular}[c]{@{}c@{}}Avg Poison \\ Success (\%) ↓\end{tabular} & \begin{tabular}[c]{@{}c@{}}Avg Natural \\ Accuracy (\%) ↑\end{tabular} & \begin{tabular}[c]{@{}c@{}}Max Poison \\ Success (\%) ↓\end{tabular} \\ \midrule
\multicolumn{1}{l|}{None}                                            & 44.00                                                     & 94.84$_{0.2}$                                                       & 43.95$_{33.6}$                                                    & 94.89$_{0.2}$                                                       & 93.59                                                         & 47.00                                                     & 93.79$_{0.2}$                                                       & 32.51$_{30.3}$                                                    & 93.76$_{0.2}$                                                       & 79.43                                                         \\
\multicolumn{1}{l|}{\epic}                       & 10.00                                                     & 85.14$_{1.2}$                                                       & 27.31$_{34.0}$                                                    & 82.20$_{1.1}$                                                       & 84.71                                                         & 27.00                                                     & 90.87$_{0.4}$                                                       & 21.53$_{28.8}$                                                    & 88.05$_{1.1}$                                                       & 80.75                                                         \\
\multicolumn{1}{l|}{\friends}                      & \textbf{0.00}                                             & \textbf{91.15$_{0.4}$}                                              & 8.32$_{22.3}$                                                     & 91.01$_{0.4}$                                                       & 83.03                                                         & \textbf{1.00}                                             & 90.09$_{0.4}$                                                       & \textbf{1.37$_{0.9}$}                                             & 90.01$_{0.2}$                                                       & 3.18                                                          \\
\multicolumn{1}{l|}{\textbf{\pebm}}                   & \textbf{0.00}                                             & \textbf{92.26$_{0.2}$}                                              & \textbf{1.27$_{0.6}$}                                             & \textbf{92.91$_{0.2}$}                                              & \textbf{2.16}                                                 & \textbf{1.00}                                             & \textbf{91.36$_{0.3}$}                                              & \textbf{1.46$_{0.8}$}                                             & \textbf{91.83$_{0.3}$}                                              & \textbf{2.49}                                                 \\
\multicolumn{1}{l|}{\textbf{\pebm-P}}                 & NA                                                          & NA                                                                & \textbf{1.38$_{0.7}$}                                             & \textbf{92.70$_{0.2}$}                                             & \textbf{2.78}                                                 & NA                                                          & NA                                                                & \textbf{1.63$_{1.0}$}                                             & \textbf{91.49$_{0.3}$}                                              & 3.47                                                          \\
\multicolumn{1}{l|}{\textbf{\pebm$_{CN-10}$}}         & \textbf{0.00}                                             & \textbf{92.99$_{0.2}$}                                              & \textbf{1.43$_{0.8}$}                                             & \textbf{92.90$_{0.2}$}                                              & \textbf{3.06}                                                 & \textbf{1.00}                                             & \textbf{92.02$_{0.2}$}                                              & \textbf{1.50$_{0.9}$}                                             & \textbf{92.03$_{0.2}$}                                              & \textbf{2.52}                                                 \\
\multicolumn{1}{l|}{\textbf{\pebm$_{IN}$}} & 1.00                                                      & \textbf{92.98$_{0.2}$}                                              & \textbf{1.39$_{0.8}$}                                             & \textbf{92.92$_{0.2}$}                                              & \textbf{2.50}                                                 & \textbf{1.00}                                             & \textbf{92.02$_{0.2}$}                                              & \textbf{1.52$_{0.8}$}                                             & \textbf{92.02$_{0.3}$}                                              & \textbf{2.81}                                                 \\
\multicolumn{1}{l|}{\textbf{\pebm-P$_{CN-10}$}}       & NA                                                          & NA                                                                & \textbf{1.64$_{0.01}$}                                            & \textbf{92.86$_{0.20}$}                                             & \textbf{4.34}                                                 & NA                                                          & NA                                                                & \textbf{1.68$_{1.0}$}                                             & \textbf{92.07$_{0.2}$}                                              & 3.34                                                          \\ \bottomrule
\end{tabular}
}
    % Second Table
    \scalebox{0.66}{% Please add the following required packages to your document preamble:
\begin{tabular}{@{}rccccccccc@{}}
\toprule
\multicolumn{10}{c}{\textbf{Transfer Learning}}                                                                                                                                                                                                                                                                                                                                                                                                                                                                                                                                                                                                                           \\ \midrule
\textbf{}                                                   & \multicolumn{5}{c}{\textbf{Fine-Tune}}                                                                                                                                                                                                                                                                                                  & \multicolumn{4}{c}{\textbf{Linear - Bullseye Polytope}}                                                                                                                                                                                                           \\ \cmidrule(lr){2-6} \cmidrule(lr){7-10}
\textbf{}                                                   & \multicolumn{2}{c}{\textbf{Bullseye Polytope-10\%}}                                                                             & \multicolumn{3}{c}{\textbf{Narcissus-10\%}}                                                                                                                                                           & \multicolumn{2}{c}{\textbf{BlackBox-10\%}}                                                                                      & \multicolumn{2}{c}{\textbf{WhiteBox-1\%}}                                                                                       \\ \cmidrule(lr){2-3} \cmidrule(lr){4-6}\cmidrule(lr){7-8} \cmidrule(lr){9-10}
\textbf{}                                                   & \begin{tabular}[c]{@{}c@{}}Poison \\ Success (\%) ↓\end{tabular} & \begin{tabular}[c]{@{}c@{}}Avg Natural \\ Accuracy (\%) ↑\end{tabular} & \begin{tabular}[c]{@{}c@{}}Avg Poison \\ Success (\%) ↓\end{tabular} & \begin{tabular}[c]{@{}c@{}}Avg Natural \\ Accuracy (\%) ↑\end{tabular} & \begin{tabular}[c]{@{}c@{}}Max Poison \\ Success (\%) ↓\end{tabular} & \begin{tabular}[c]{@{}c@{}}Poison \\ Success (\%) ↓\end{tabular} & \begin{tabular}[c]{@{}c@{}}Avg Natural \\ Accuracy (\%) ↑\end{tabular} & \begin{tabular}[c]{@{}c@{}}Poison \\ Success (\%) ↓\end{tabular} & \begin{tabular}[c]{@{}c@{}}Avg Natural \\ Accuracy (\%) ↑\end{tabular} \\ \midrule
\multicolumn{1}{l|}{None}                                   & 46.00                                                     & 89.84$_{0.9}$                                                       & 33.41$_{33.9}$                                                    & 90.14$_{2.4}$                                                       & 98.27                                                         & 93.75                                                     & 83.59$_{2.4}$                                                       & 98.00                                                     & 70.09$_{0.2}$                                                       \\
\multicolumn{1}{l|}{\epic}              & 42.00                                                     & 81.95$_{5.6}$                                                       & 20.93$_{27.1}$                                                    & 88.58$_{2.0}$                                                       & 63.00                                                     & 66.67                                                     & 84.34$_{3.8}$                                                       & 91.00                                                     & 64.79$_{0.7}$                                                       \\
\multicolumn{1}{l|}{\friends}             & 8.00                                                      & 87.82$_{1.2}$                                                       & 3.04$_{5.1}$                                                      & 89.81$_{0.5}$                                                       & 17.32                                                         & 33.33                                                     & 85.18$_{2.3}$                                                       & 19.00                                                     & 60.90$_{0.6}$                                                       \\
\multicolumn{1}{l|}{\textbf{\pebm}}          & \textbf{0.00}                                             & \textbf{88.95$_{1.1}$}                                              & \textbf{1.98$_{1.7}$}                                             & \textbf{91.40$_{0.4}$}                                              & \textbf{5.98}                                                 & \textbf{0.00}                                             & \textbf{92.89$_{0.2}$}                                              & \textbf{6.00}                                             & \textbf{64.51$_{0.6}$}                                              \\
\multicolumn{1}{l|}{\textbf{\pebm-p}}        & NA                                                          & NA                                                                & \textbf{3.66$_{4.63}$}                                                    & \textbf{90.89$_{0.31}$}                                                       & 16.04                                                         & NA                                                          & NA                                                                & NA                                                          & NA                                                                \\
\multicolumn{1}{l|}{\textbf{\pebm$_{CN-10}$}} & \textbf{0.00}                                             & \textbf{88.67$_{1.2}$}                                              & \textbf{2.97$_{2.5}$}                                             & \textbf{90.99$_{0.3}$}                                              & \textbf{7.95}                                                 & \textbf{0.00}                                             & \textbf{92.82$_{0.1}$}                                              & \textbf{6.00}                                             & \textbf{64.44$_{0.4}$}                                              \\
\multicolumn{1}{l|}{\textbf{\pebm$_{IN}$}}            & \textbf{0.00}                                             & \textbf{87.52$_{1.2}$}                                              & \textbf{2.02$_{1.0}$}                                             & \textbf{89.78$_{0.6}$}                                              & \textbf{3.85}                                                 & \textbf{0.00}                                             & \textbf{92.38$_{0.3}$}                                              & \textbf{6.00}                                             & \textbf{64.98$_{0.3}$}                                              \\ \bottomrule
\vspace{-10mm}
\end{tabular}}
    \label{table:core_results}
\end{table*}

We compare our method, \pebm, against previous state-of-the-art defenses \epic and \friends  on the current leading triggered poison, Narcissus (NS) and triggerless poisons, Gradient Matching (GM) and Bullseye Polytope (BP). Triggerless attacks GM and BP have 100 and 50 poison scenarios while NS has 10 (one per class). Primary results use a ResNet18 classifier and the CIFAR-10 dataset. We train a variety of EBMs using the training techniques described in App. \ref{sec:ebm_1} with specific datasets for our experimental results:
% $\mathcal{X} \setminus \bigcup_{j=1}^{J} \Pi_j$
\begin{enumerate}
    \itemsep0em 
    \item \textbf{\pebm}: To ensure EBM training is blind to poisoned images, we exclude the indices for all potential poison scenarios which resulted in 37k, 45k, and 48k training samples for GM, NS, and BP respectively of the original 50k CIFAR-10 train images. 
    
    % In this situation, we assume that we have oracle knowledge of which points are being poisoned to mimic a clean In-Distribution dataset, but no knowledge of the poisons or training paradigms themselves.

    \item \textbf{\pebm-P}: Trained on the full CIFAR-10 dataset in which 100\% of training samples are poisoned using their respective class' NS poison trigger. This model explores the ability to learn robust features even when the EBM is exposed to full adversarial influences during training (even beyond the strongest classifier scenario of 10\% poison).

    \item \textbf{\pebm$_{\text{CN-10}}$}: Trained on the CINIC-10 dataset, which is a mix of ImageNet (70k) and CIFAR-10 (20k) images where potential poison samples are removed from CIFAR-10 indices \cite{darlow2018cinic10}. This model investigates the effectiveness of EBM purification when trained on a distributionally similar dataset. 
    % \om{how do u know that it has a similar distribution?} \om{maybe contains the same classes?}

    \item \textbf{\pebm$_{\text{IN}}$}: Trained exclusively on the ImageNet (70k) portion of the CINIC-10 dataset. This model tests the generalizability of the EBM purification process on a public out-of-distribution (POOD) dataset that shares no direct overlap with the classifier's training data $\mathcal{X}$.

    \item \textbf{\pebm-P$_{\text{CN-10}}$}: Trained on the CINIC-10 dataset where the CIFAR-10 subset is fully poisoned. This variant examines the EBM's ability to learn and purify data where a significant portion of the training dataset is adversarially manipulated and the clean images are from a POOD dataset. 
\end{enumerate}

A single hyperparameter grid-search for Langevin dynamics was done on the \pebm model using a single poison scenario per training paradigm (from scratch, transfer linear and transfer fine-tune) as seen in App. \ref{app:grid_search}. The percentage of classifier training data poisoned is indicated next to each poison scenario. Additional details on poison sources, poison crafting, definitions of poison success, and training hyperparameters can be found in App. \ref{app:poison_implement}.

\subsection{Benchmark Results}

Table \ref{table:core_results} shows our primary results in which \textbf{\pebm achieves state-of-the-art (SoTA) poison defense and natural accuracy in all poison scenarios} and fully poisoned \pebm-P achieves SoTA performance for Narcissus. 
Furthermore, \textbf{ all public out-of-distribution (POOD) EBMs achieve SoTA performance in almost every category} without additional hyperparameter search. 

For GM, \pebm matches SoTA in a nearly complete poison defense and achieves 1.1\% less natural accuracy degradation, from no defense, than the previous SoTA. For BP, \pebm exceeds the previous SoTA with an 8-33\% poison defense reduction and 1.1-7.5\% less degradation in natural accuracy. For NS, \pebm matches or exceeds previous SoTA with a 1-8\% poison defense reduction and ~1.5\% less degradation in natural accuracy.

\subsection{Results on Additional Models and Datasets}

Table \ref{table:cinic_results} shows results when we apply NS poisons (generated using CIFAR-10) to the CINIC-10 dataset. To ensure no overlap for our EBMs, we train on CINIC-10's validation set, which has the same size and composition as its training set. Table \ref{table:model_results} shows results for MobileNetV2 and DenseNet121 architectures. \textbf{\pebm is SoTA across all models and in CINIC-10 NS poison scenarios} showing no performance dependence on dataset or model. Full results are in App. \ref{app:results}.

Finally, the Hyperlight Benchmark CIFAR-10 (HLB) is a drastically different case study from our standard benchmarks with a residual-less network architecture, unique initialization scheme, and super-convergence training method that recently held the world record of achieving 94\% test accuracy on CIFAR-10 using a surprising total of 10 epochs \cite{Balsam2023hlbCIFAR10}. We observe that NS still successfully poisons the HLB model, and does so by the end of the first epoch. Applying \epic and \friends becomes unclear, as they use model information after a warm-up period, but we choose the most sensible warm-up period of one epoch, even though the poisons have set in. From Table \ref{table:model_results} subset selection based \epic is unable to train effectively, and \friends offers some defense. \pebm still applies with minimal adjustment to the training pipeline and defends effectively against these poisons. Table \ref{table:model_results} also shows the effect of differing MCMC steps where 25 MCMC steps already offers comparable defense to \friends, and by 50 steps, \pebm shows SoTA poison defense and natural accuracy. Increasing steps further reduces poison success, but at the cost of natural accuracy and linearly increasing preprocessing time. 

The last column of the HLB section shows timing analysis on a NVIDIA A100 GPU. Due to HLB training speeds, timings primarily indicate the processing time of the defenses. 
\pebm is faster in total train time and per epoch time than existing SoTA defense methods. We emphasize that, in practice, \pebm can be applied once to a dataset and used across model architectures, unlike previous SoTA defenses \epic and \friends, which require train-time information on model outputs. See App. \ref{app:timing} for further timing. 

\begin{table}[ht]
   % \vspace{-1mm}
   \centering
   \footnotesize
   \caption{Poison success and natural accuracy when training on CINIC-10 Dataset From Scratch Results with NS Poison}
   \vspace{-2mm}
   \label{table:cinic_results}
   \scalebox{0.68}{% Please add the following required packages to your document preamble:
% \usepackage{booktabs}
% \usepackage{booktabs}
\begin{tabular}{@{}lcccc@{}}
\toprule
\multicolumn{5}{c}{\textbf{CINIC-10 Narcissus - 1\% From-Scratch (200 Epochs)}}                                                                                                                                                                                                                                                                                                                           \\ \midrule
                                             & \multicolumn{1}{c}{\begin{tabular}[c]{@{}c@{}}Avg Poison \\ Success (\%) ↓\end{tabular}} & \multicolumn{1}{c}{\begin{tabular}[c]{@{}c@{}}Avg Natural\\  Accuracy (\%) ↑\end{tabular}} & \multicolumn{1}{c}{\begin{tabular}[c]{@{}c@{}}Max Poison \\ Success (\%) ↓\end{tabular}} & \multicolumn{1}{c}{\begin{tabular}[c]{@{}c@{}}CIFAR-10\\ Accuracy (\%) ↑\end{tabular}} \\ \midrule
\multicolumn{1}{l|}{None}                    & 62.06$_{0.21}$                                                                      & 86.32$_{0.10}$                                                                          & 90.79                                                                           & 94.22$_{0.16}$                                                                      \\
\multicolumn{1}{l|}{\epic}    & 49.50$_{0.27}$                                                                        & 81.91$_{0.08}$                                                                          & 91.35                                                                           & 91.10$_{0.21}$                                                                      \\
\multicolumn{1}{l|}{\friends} & 11.17$_{0.25}$                                                                        & 77.53$_{0.60}$                                                                          & 82.21                                                                           & 88.27$_{0.68}$                                                                      \\
\multicolumn{1}{l|}{\pebm}    & \textbf{7.73$_{0.08}$}                                                                & \textbf{82.37$_{0.14}$}                                                               & \textbf{29.48}                                                                  & \textbf{91.98$_{0.16}$}                                                           \\ \bottomrule
\vspace{-6mm}
\end{tabular}}
\end{table}

\begin{table}[ht]
    \vspace{-0.5mm}
   \centering
   \footnotesize
   \caption{MobileNetV2 and DenseNet121 results and HyperlightBench for a novel training paradigm where \pebm is still effective.}
   \vspace{-2mm}
   \label{table:model_results}
   \scalebox{0.68}{
\begin{tabular}{@{}rcccc@{}}
\toprule
                                        \multicolumn{5}{c}{\textbf{From Scratch NS-1\% (200 epochs)}}                                                                                                                                                                                                                     \\ \midrule
\textbf{}                                              & \multicolumn{2}{c}{\textbf{MobileNetV2}}                                                                                                   & \multicolumn{2}{c}{\textbf{DenseNet121}}                                                                                             \\ \cmidrule(lr){2-3}\cmidrule(lr){4-5} 
\textbf{}                                              & \begin{tabular}[c]{@{}c@{}}Avg Poison \\ Success  (\%) ↓\end{tabular} & \begin{tabular}[c]{@{}c@{}}Avg Natural \\ Accuracy (\%) ↑\end{tabular}       & \begin{tabular}[c]{@{}c@{}}Avg Poison \\ Success  (\%) ↓\end{tabular} & \begin{tabular}[c]{@{}c@{}}Avg Natural \\ Accuracy (\%) ↑\end{tabular} \\ \midrule
\multicolumn{1}{l|}{None}                              & 32.70$_{0.25}$                                                     & 93.92$_{0.13}$                                             & 46.52$_{32.2}$                                                                        & 95.33$_{0.1}$                                                                                                             \\
\multicolumn{1}{l|}{\epic}              & 22.35$_{0.24}$                                                     & 78.16$_{9.93}$                                                            & 32.60$_{29.4}$                                                                        & 85.12$_{2.4}$                                                         \\
\multicolumn{1}{l|}{\friends}           & 2.00$_{0.01}$                                                      & 88.82$_{0.57}$                                                             & 8.60$_{21.2}$                                                                         & 91.55$_{0.3}$                                                           \\
\multicolumn{1}{l|}{\textbf{\pebm}}     & \textbf{1.64$_{0.01}$}                                             & \textbf{91.75$_{0.13}$}                                                   & \textbf{1.42$_{0.7}$}                                                                 & \textbf{93.48$_{0.1}$}                                                         \\ \midrule
                                      \multicolumn{5}{c}{\textbf{Linear Transfer  WhiteBox BP-10\%}}                                                                                                                                                                                                                    \\ \midrule
                                                       & \multicolumn{2}{c}{\textbf{MobileNetV2}}                                                                                                   & \multicolumn{2}{c}{\textbf{DenseNet121}}                                                                                             \\ \cmidrule(lr){2-3}\cmidrule(lr){4-5} 
                                                       & \begin{tabular}[c]{@{}c@{}}Poison \\ Success (\%) ↓\end{tabular}      & \begin{tabular}[c]{@{}c@{}}Avg Natural \\ Accuracy (\%) ↑\end{tabular}       & \begin{tabular}[c]{@{}c@{}}Poison \\ Success (\%) ↓\end{tabular}      & \begin{tabular}[c]{@{}c@{}}Avg Natural \\ Accuracy (\%) ↑\end{tabular} \\ \midrule
\multicolumn{1}{l|}{None}                              & 81.25                                                          & 73.27$_{0.97}$                                                            & 73.47                                                          & 82.13$_{1.62}$                                                      \\
\multicolumn{1}{l|}{\epic}              & 56.25                                                          & 54.47$_{5.57}$                                                            & 41.67                                                     & 70.13$_{5.2}$                                                     \\
\multicolumn{1}{l|}{\friends}           & 41.67                                                          & 68.86$_{1.50}$                                                            & 56.25                                                     & 80.12$_{1.8}$                                                      \\
\multicolumn{1}{l|}{\textbf{\pebm}}     & \textbf{0.00}                                                  & \textbf{78.57$_{1.37}$}                                                   & \textbf{0.00}                                                  & \textbf{89.29$_{0.94}$}                                             \\ \midrule
      \multicolumn{5}{c}{\textbf{Hyperlight Bench CIFAR-10 NS-1\% (10 Epochs)}}                                                                                                                                                                                                         \\ \midrule
                                                       & \begin{tabular}[c]{@{}c@{}}Avg Poison \\ Success (\%) ↓\end{tabular} & \begin{tabular}[c]{@{}c@{}}Avg Natural \\ Accuracy (\%) ↑\end{tabular} & \begin{tabular}[c]{@{}c@{}}Max Poison\\ Success (\%) ↓\end{tabular}   & \begin{tabular}[c]{@{}c@{}}Train Time \\ (s)\end{tabular} \\ \midrule
\multicolumn{1}{l|}{None}                              & 76.39$_{16.35}$                                                  & 93.95$_{0.10}$                                                          & 95.69                                                            & 6.81$_{0.62}$                                                                  \\
\multicolumn{1}{l|}{\epic}              & 10.58$_{18.35}$                                                  & 24.88$_{6.04}$                                                          & 50.21                                                            & 612.43$_{30.16}$                                                             \\
\multicolumn{1}{l|}{\friends}           & 11.35$_{18.45}$                                                  & 87.03$_{1.52}$	                                                          & 56.65	                                                            & 427.50$_{0.50}$                                                                \\
\multicolumn{1}{l|}{\textbf{\pebm-25}}  & 10.59$_{26.04}$                                                  & 92.75$_{0.13}$                                                          & 84.60                                                            & 54.70$_{0.48}$                                                               \\
\multicolumn{1}{l|}{\textbf{\pebm-50}}  & \textbf{2.16$_{1.22}$}                                           & \textbf{92.38$_{0.17}$}                                                 & \textbf{3.74}                                                    & 92.89$_{0.48}$                                                               \\
\multicolumn{1}{l|}{\textbf{\pebm-100}} & \textbf{1.89$_{1.06}$}                                           & \textbf{91.94$_{0.14}$}                                                 & \textbf{3.47}                                                    & 168.69$_{0.46}$                                                              \\
\multicolumn{1}{l|}{\textbf{\pebm-150}} & \textbf{1.93$_{1.15}$}                                           & \textbf{91.46$_{0.17}$}                                                 & \textbf{4.14}                                                    & \multicolumn{1}{c}{244.72$_{0.47}$ }                                            \\
\multicolumn{1}{l|}{\textbf{\pebm-300}} & \textbf{1.68$_{0.82}$}                                           & \textbf{90.55$_{0.21}$}                                                 & \textbf{2.89}                                                    & \multicolumn{1}{c}{478.29$_{0.47}$ }                                            \\ \bottomrule
\vspace{-7mm}
\end{tabular}}
\end{table}

\subsection{Further Experiments}

\begin{figure}[ht]
    \vspace{-2mm}
    \centering
    \includegraphics[width=\linewidth]{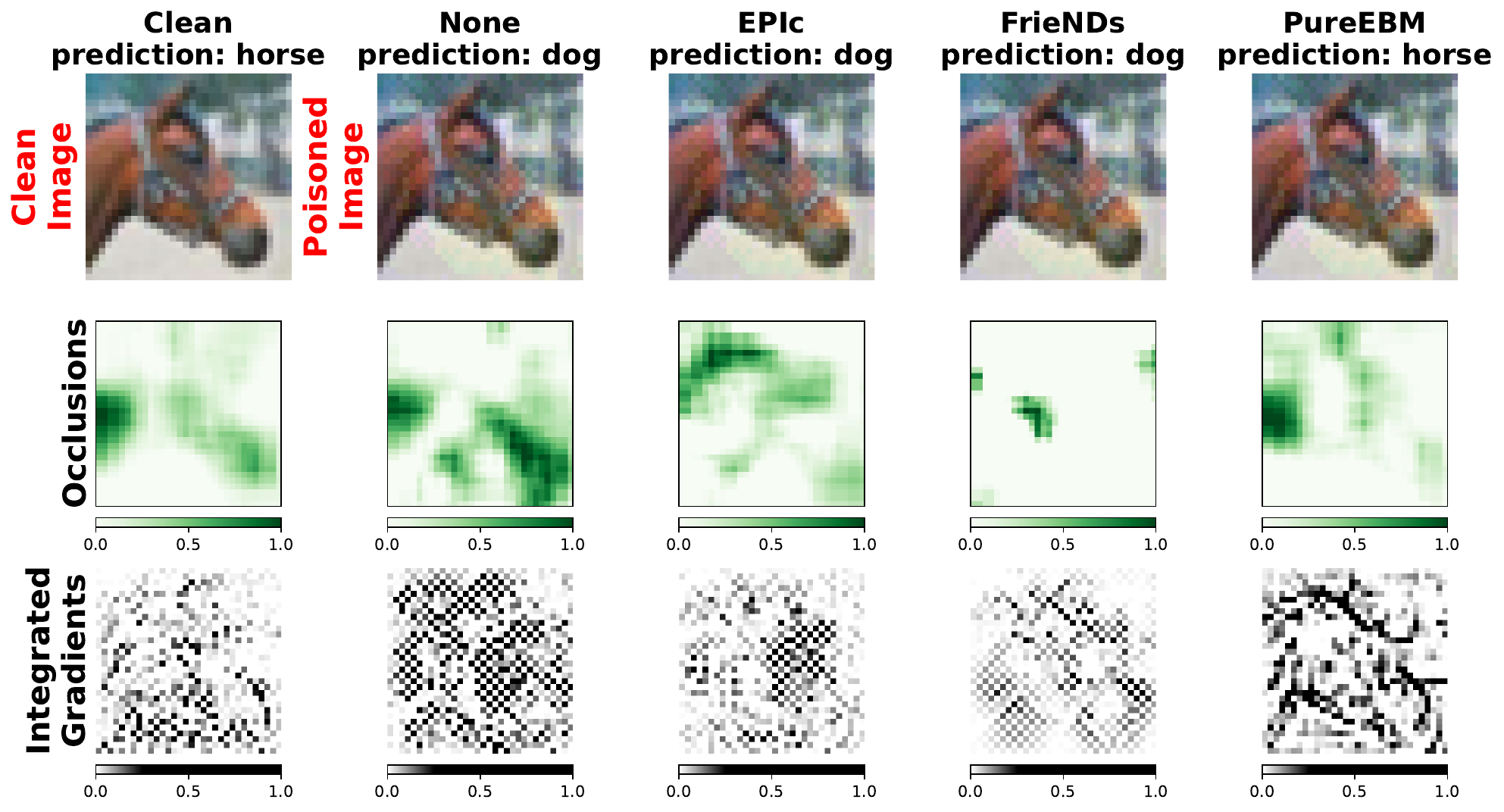}
    \vspace{-4mm}
    \caption{Defense Interpretability: Model using \pebm focuses on the outline of the horse in the occlusions analysis and to a higher degree on the primary features in the gradient space than even the clean model on clean data.}
    \label{fig:captum}
    % \vspace{-2mm}
\end{figure}

\textbf{Model Interpretability} Using the Captum interpretability library, in Figure \ref{fig:captum}, we compare a clean model with clean data to the various defense techniques on a sample image poisoned with the NS Class 5 trigger $\rho$ \cite{kokhlikyan2020captum}. Only the clean model and the model that uses \pebm  correctly classify the sample as a horse, and the regions most important to prediction, via occlusion analysis, most resemble the shape of a horse in the clean and \pebm images. Integrated Gradient plots show how \pebm actually enhances interpretability of relevant features in the gradient space for prediction compared to even the clean NN. Aditionally we see that the NN trained with \pebm is less sensitive to input perturbations compared to all other NNs. See App. \ref{app:captum} for additional examples.

\textbf{Flatter solutions are robust to Poisons} Recently \citet{liu2023friendly} showed that effective poisons introduce a local sharp region with a high training loss and that an effective defense can smooth the loss landscape of the classifier. We consider the curvature of the loss with respect to our model's weights as a way to evaluate defense success. The PSGD framework \citep{psgd,psgd_affine,li2022black,pooladzandi2024curvatureinformed} estimates the Hessian of the loss $\mathbf{H}$ of the model over the full dataset and the poisoned points through training. In information theory, $0.5\log\det(\mathbf{H})$ is a good proxy for the description length of the model parameters. We find that training with data points pre-processed by the \pebm stochastic transformation $\Psi_{T}(x)$ reduces the curvature of the loss of the NN over the full dataset and around poisoned points. In effect, NNs trained with points defended with \pebm are significantly more robust to perturbation than other defenses. In App. \ref{app:poi-diverge}, we find that \pebm and \friends models' parameters diverge from poisoned models more so than \epic. 

% Interestingly, we see that \friends and \epic result in NNs with similar curvature. 
 % Using \om{psgd} to estimate the curvature of models, we estimate how flat, or robust to perturbation, of a solution each defense method finds by taking the negative Log-Determinant of the the Hessian approximation $\log\left(\left|\text{det}(\mathbf{H})\right|\right)$, where larger negative value represents a flatter, more robust solution basin in the parameter space. This estimation is done using the average of 8 models (first 8 GM target indices) trained in the 80 epoch pipeline, and then further trained for 3 epochs using the 2nd order optimizer in which the Hessian approximation is utilized \cite{psgd}. In Figure \ref{fig:log_det} pureEBM is able to find the flatter, more robust network weights of the defense methods against the Gradient Matching poison when training on both the full dataset with poisoned samples or just the poisoned subset of data. 

\begin{figure}[ht]
    \vspace{-2mm}
    \centering
    \includegraphics[width=\linewidth]{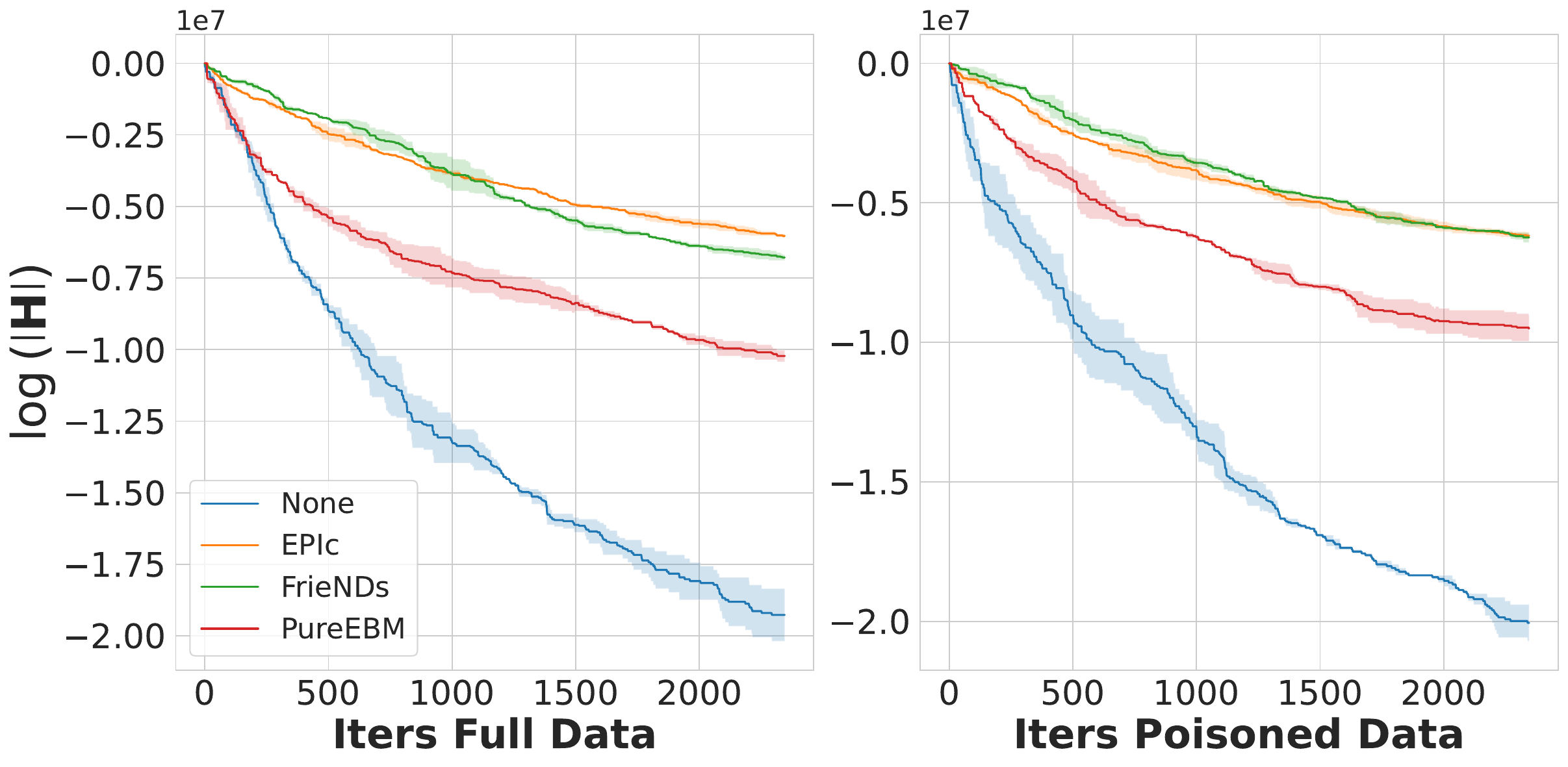}
    \caption{Estimate loss curvature - classifier robustness - with $\log\left(\left|\mathbf{H}\right|\right)$ against both full and poisoned subset of training data. Model trained with \pebm has the lowest curvature compared to SoTA defense methods.}
    \label{fig:log_det}
    \vspace{-2mm}
\end{figure}

\vspace{-2mm}
\section{Conclusion}
Poisoning has the potential to become one of the greatest attack vectors to AI models, decreasing model security and eroding public trust. Further discussion of ethics and impact can be found in App. \ref{app:impact}. In this work, we present \pebm, a powerful Energy-Based Model defense against imperceptible train time data poisoning attacks. Our approach significantly advances the field of poison defense and model security by addressing the critical challenge of adversarial poisons in a manner that maintains high natural accuracy and method generality. Through extensive experimentation, \pebm has demonstrated state-of-the-art performance in defending against a range of poisoning scenarios using the leading Gradient Matching, Narcissus, and Bullseye Polytope attacks. The key to our method's success is a stochastic preprocssing step that uses MCMC sampling with an EBM to iteratively purify poisoned samples, moving them into a lower energy, natural data manifold. We share similar SoTA results with EBMs trained on out-of-distribution and poisoned datasets, underscoring the method's adaptability and robustness. A versatile, efficient, and robust method for purifying training data, \pebm sets a new standard in the ongoing effort to fortify machine learning models against the evolving threat of data poisoning attacks. Because \pebm neutralizes all SoTA data poisoning attacks effectively, we believe our research can have a significant \textbf{positive social impact} to inspire trust in widespread machine learning adoption.

\section{Acknowledgments}
This work is supported with Cloud TPUs from Google’s Tensorflow Research Cloud (TFRC). We would like to acknowledge Jonathan Mitchell, Mitch Hill, Yuan Du and Kathrine Abreu for support on base EBM code. As well as a Xi-Lin Li for his insight of collecting curvature information to see if training on samples from \pebm give a solution that is more robust to input perturbations compared to other defenses. And Yunzheng Zhu for his help in crafting poisons. An early version of this work was originally published in author Jeffrey Jiang's thesis \cite{Jiang2024-cr}.

\bibliography{ms}
\bibliographystyle{icml2023}

%%%%%%%%%%%%%%%%%%%%%%%%%%%%%%%%%%%%%%%%%%%%%%%%%%%%%%%%%%%%%%%%%%%%%%%%%%%%%%%
%%%%%%%%%%%%%%%%%%%%%%%%%%%%%%%%%%%%%%%%%%%%%%%%%%%%%%%%%%%%%%%%%%%%%%%%%%%%%%%
% APPENDIX
%%%%%%%%%%%%%%%%%%%%%%%%%%%%%%%%%%%%%%%%%%%%%%%%%%%%%%%%%%%%%%%%%%%%%%%%%%%%%%%
%%%%%%%%%%%%%%%%%%%%%%%%%%%%%%%%%%%%%%%%%%%%%%%%%%%%%%%%%%%%%%%%%%%%%%%%%%%%%%%
\newpage
\appendix
\onecolumn

\section{EBM Further Background}
\subsection{Chaotic Dynamics}\label{sec:chaos}
Chaos theory offers a distinct perspective for justifying the suppression of adversarial signals through extended iterative transformations. In deterministic systems, chaos is characterized by the exponential growth of initial infinitesimal perturbations over time, leading to a divergence in the trajectories of closely situated points — a phenomenon popularly known as the butterfly effect. This concept extends seamlessly to stochastic systems as well. \citet{hill2021stochastic} were the first to show the chaotic nature of EBMs for purification. Here we verify that both poisoned images and clean images have the same chaotic properties.

\subsubsection*{Stochastic Differential Equations and Chaos}
Consider the Stochastic Differential Equation (SDE) given by:
\begin{align}
\text{d}X_t = V(X)\text{dt}+ {\eta_{noise}} \text{d}B_t, \label{eqn:sde}
\end{align}
where $B_t$ denotes Brownian motion and ${\eta_{noise}} \ge 0$. This equation, which encompasses the Langevin dynamics, is known to exhibit chaotic behavior in numerous contexts, especially for large values of ${\eta_{noise}}$ \cite{lai2003unstable}.

\subsubsection*{Maximal Lyapunov Exponent}
The degree of chaos in a dynamical system can be quantified by the maximal Lyapunov exponent $\lambda$, defined as:
\begin{align}
\lambda = \lim_{t \rightarrow \infty} \frac{1}{t} \log \frac{|\delta X_{{\eta_{noise}}}(t)|}{|\delta X_{{\eta_{noise}}}(0)|}, \label{eqn:lyapunov}
\end{align}
where $\delta X_{\eta_{noise}} (t)$ represents an infinitesimal perturbation in the system state at time $t$, evolved according to Eq. \ref{eqn:sde} from an initial perturbation $\delta X_{\eta_{noise}} (0)$. For ergodic dynamics, $\lambda$ is independent of the initial perturbation $\delta X_{\eta_{noise}} (0)$. An ordered system exhibits a maximal Lyapunov exponent that is non-positive, while chaotic systems are characterized by a positive $\lambda$. Thus, by analyzing the maximal Lyapunov exponent of the Langevin equation, one can discern whether the dynamics are ordered or chaotic.

Following the classical approach outlined by \citet{benettin1976entropy}, we calculate the maximal Lyapunov exponent for the modified Langevin transformation, described by the equation:
\begin{equation}
Z_{\eta_{noise}} (X)  = x_{\tau} - \Delta \tau \nabla_{x_{\tau}} \mathcal{G}_{\theta}(x_{\tau}) + 
\eta_{noise} \sqrt{2\Delta \tau} \epsilon_{\tau}  ,
\end{equation}

This computation is performed across a range of noise strengths $\eta_{noise}$. 
\begin{figure}[ht] %[H]
    \centering
    \includegraphics[width=.6\textwidth]{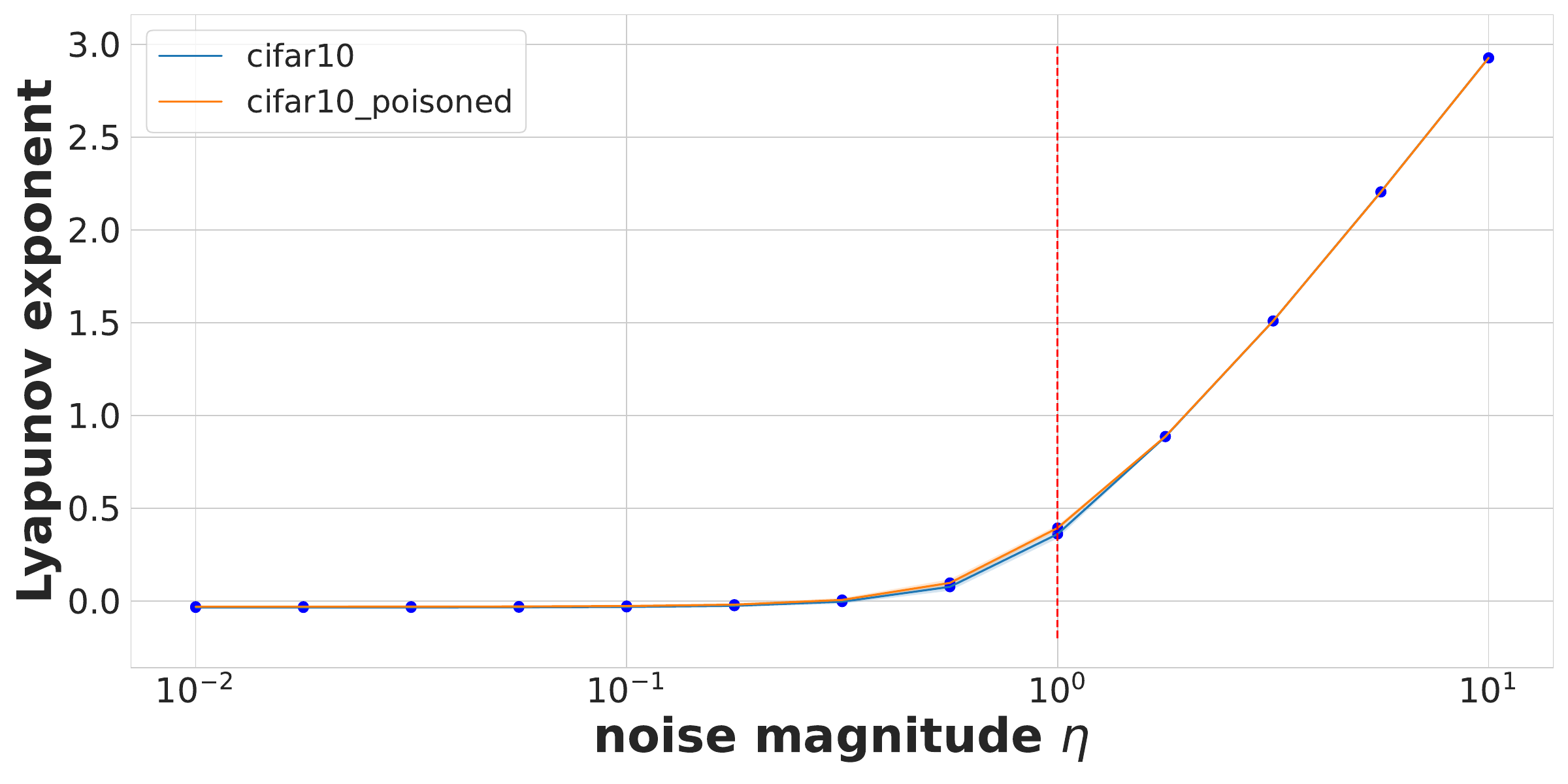} \; \includegraphics[width=.2\textwidth]{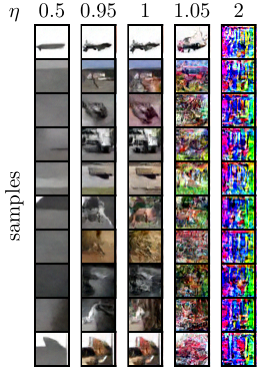}
    \caption{\emph{Left:} The maximal Lyapunov exponent varies significantly with different values of the noise parameter $\eta_{noise}$. Notably, at $\eta_{noise}= 1$, which is the setting used in our training and defense dynamics, there is a critical transition observed. This transition is from an ordered region, where the maximal exponent is zero, to a chaotic region characterized by a positive maximal exponent. This observation is crucial for understanding the underlying dynamics of our model. \emph{Right:} The appearance of steady-state samples exhibits marked differences across the spectrum of $\eta_{noise}$ values. For lower values of $\eta_{noise}$, the generated images tend to be oversaturated. Conversely, higher values of $\eta_{noise}$ result in noisy images. However, there exists a narrow window around $\eta_{noise} = 1$ where a balance is achieved between gradient and noise forces, leading to realistic synthesis of images.}
    \label{fig:lya}
\end{figure}
Our findings demonstrate a clear transition from noise-dominated to chaos-dominated behavior. Notably, at $\eta_{noise} = 1$ — the parameter setting for our training and defense algorithms — the system transitions from ordered to chaotic dynamics. This critical interval balances the ordered gradient forces, which encourage pattern formation, against chaotic noise forces that disrupt these patterns. Oversaturation occurs when the gradient forces prevail, leading to noisy images when noise is dominant. These results are illustrated in Figure~\ref{fig:lya}.

The inherent unpredictability in the paths under $Z_{\eta_{noise}}$ serves as an effective defense mechanism against targeted poison attacks. Due to the chaotic nature of the transformation, generating informative attack gradients that can make it through the defense while causing a backdoor in the network becomes challenging.  Exploring other chaotic transformations, both stochastic and deterministic, could be a promising direction for developing new defense strategies.

We see that as expected the Lyapunov exponent of the Langevin dynamics on clean and poisoned points are exactly the same. 
\subsection{EBM Purification is a Convergent Process}
\label{app:convergent_EBM}

\begin{figure}
    \centering
    \includegraphics[width=0.5\linewidth]{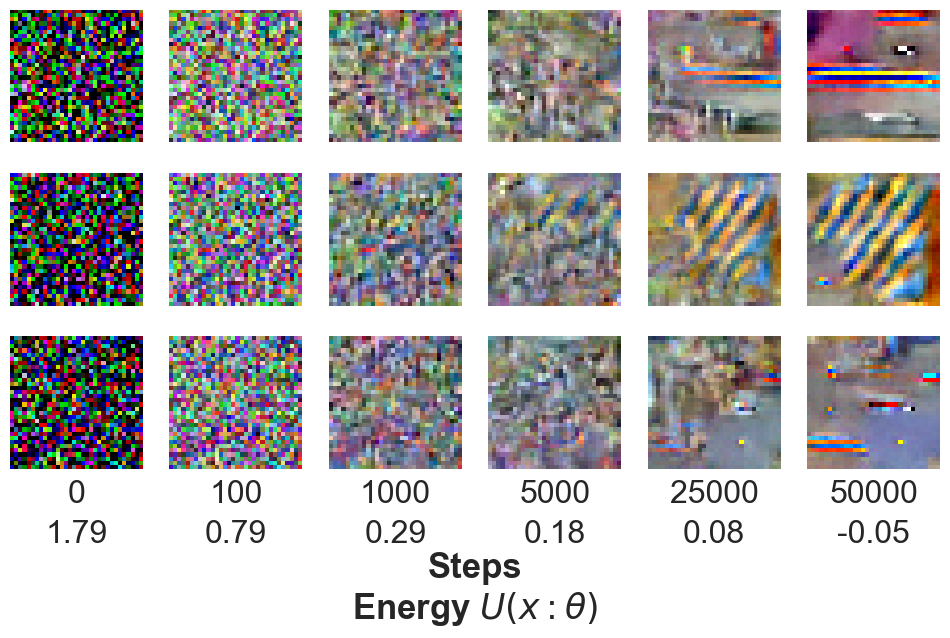}
    \caption{Random Noise initialization of purification process}
    \label{fig:rand_noise_images}
\end{figure}

Energy-based models and Langevin dynamics are both commonly associated with divergent generative models and diffusion processes in the machine learning community, in which samples are generated from a random initialization using a conditional or unconditional probability distribution. In contrast, we emphasize that the EBM and MCMC purification process is a convergent generative chain, initialized with a sample from some data distribution $p_{data}$ with metastable properties that retain features of the original image due to the low energy density around the image \cite{nijkamp2019anatomy}. To illustrate this point, Figure \ref{fig:rand_noise_images} shows the purification process on random noise initialization. Even with long-run dynamics of 50k Langevin steps producing low energy outputs, the resulting `images' are not meaningful, highlighting the desired reliance on a realistic sample initializing a convergent MCMC chain. Previous analysis demonstrates the mid-run memoryless properties that remove adversarial poisons and enable the EBM purification process once paired with the metastable aspects of the convergent MCMC chain. 

\newpage
\section{Additional Results}\label{app:results}

\subsection{Full Results Primary Experiments}\label{app:core_full_results}

Results on all primary poison scenarios with ResNet18 classifier including all \epic versions (various subset sizes and selection frequency), \friends versions (bernouilli or gaussian added noise trasnform), and all natural \pebm versions. Asterisk (*) indicates a baseline defense that was selected for the main paper results table due to best poison defense performance. 

We note that the implemention made available for \epic contains discrepancies, occasionally returning random subsets, and drops repeatedly selected points every epoch. We did our best to reproduce results, and choose the best of all version ran to compare to. Further, we note that our results outperform the results reported by \citet{yang2022poisons}, listed in the table here as \epic$_{reported}$.

\begin{table}[H]
    \centering
    % \caption{Experimental Results for all primary poison scenarios}
    % First Table
    \scalebox{0.6}{% Please add the following required packages to your document preamble:
\begin{tabular}{@{}rcccccccccc@{}}
\toprule
\multicolumn{11}{c}{\textbf{From Scratch}}                                                                                                                                                                                                                                                                                                                                                                                                                                                                                                                                                                                                                                                                                                               \\ \midrule
\textbf{}                                                            & \multicolumn{5}{c}{\textbf{200 - Epochs}}                                                                                                                                                                                                                                                                                               & \multicolumn{5}{c}{\textbf{80 - Epochs}}                                                                                                                                                                                                                                                                                                \\ \cmidrule(lr){2-6} \cmidrule(lr){7-11} 
\textbf{}                                                            & \multicolumn{2}{c}{\textbf{Gradient Matching-1\%}}                                                                              & \multicolumn{3}{c}{\textbf{Narcissus-1\%}}                                                                                                                                                            & \multicolumn{2}{c}{\textbf{Gradient Matching-1\%}}                                                                              & \multicolumn{3}{c}{\textbf{Narcissus-1\%}}                                                                                                                                                            \\ \cmidrule(lr){2-3} \cmidrule(lr){4-6} \cmidrule(lr){7-8} \cmidrule(lr){9-11} 
\textbf{}                                                            & \begin{tabular}[c]{@{}c@{}}Poison \\ Success (\%) ↓\end{tabular} & \begin{tabular}[c]{@{}c@{}}Avg Natural \\ Accuracy (\%) ↑\end{tabular} & \begin{tabular}[c]{@{}c@{}}Avg Poison \\ Success (\%) ↓\end{tabular} & \begin{tabular}[c]{@{}c@{}}Avg Natural \\ Accuracy (\%) ↑\end{tabular} & \begin{tabular}[c]{@{}c@{}}Max Poison \\ Success (\%) ↓\end{tabular} & \begin{tabular}[c]{@{}c@{}}Poison \\ Success (\%) ↓\end{tabular} & \begin{tabular}[c]{@{}c@{}}Avg Natural \\ Accuracy (\%) ↑\end{tabular} & \begin{tabular}[c]{@{}c@{}}Avg Poison \\ Success (\%) ↓\end{tabular} & \begin{tabular}[c]{@{}c@{}}Avg Natural \\ Accuracy (\%) ↑\end{tabular} & \begin{tabular}[c]{@{}c@{}}Max Poison \\ Success (\%) ↓\end{tabular} \\ \midrule
\multicolumn{1}{l|}{None}                                            & 44.00                                                     & 94.84$_{0.2}$                                                       & 43.95$_{33.6}$                                                    & 94.89$_{0.2}$                                                       & 93.59                                                         & 47.00                                                     & 93.79$_{0.2}$                                                       & 32.51$_{30.3}$                                                    & 93.76$_{0.2}$                                                       & 79.43                                                         \\
\multicolumn{1}{l|}{\epic-0.1*}                       & 34.00                                                     & 91.27$_{0.4}$                                                       & 30.18$_{32.2}$                                                    & 91.17$_{0.2}$                                                       & 81.50                                                         & 27.00                                                     & 90.87$_{0.4}$                                                       & 24.15$_{30.1}$                                                    & 90.92$_{0.4}$                                                       & 79.42                                                         \\
\multicolumn{1}{l|}{\epic-0.2}                        & 21.00                                                     & 88.04$_{0.7}$                                                       & 32.50$_{33.5}$                                                    & 86.89$_{0.5}$                                                       & 84.39                                                         & 28.00                                                     & 91.02$_{0.4}$                                                       & 23.75$_{29.2}$                                                    & 89.72$_{0.3}$                                                       & 74.28                                                         \\
\multicolumn{1}{l|}{\epic-0.3*}                       & 10.00                                                     & 85.14$_{1.2}$                                                       & 27.31$_{34.0}$                                                    & 82.20$_{1.1}$                                                       & 84.71                                                         & 44.00                                                     & 92.46$_{0.3}$                                                       & 21.53$_{28.8}$                                                    & 88.05$_{1.1}$                                                       & 80.75                                                         \\
\multicolumn{1}{l|}{\epic$_{reported}$}                       & 1.00                                                     & 90.26                                                       & NA                                                    & NA                                                      & NA                                                         & NA                                                     & NA                                                       & NA                                                    & NA                                                       & NA                                                        \\
\multicolumn{1}{l|}{\friends-B}                       & 1.00                                                      & \textbf{91.16$_{0.4}$}                                              & 8.32$_{22.3}$                                                     & 91.01$_{0.4}$                                                       & 71.76                                                         & 2.00                                                      & 90.07$_{0.4}$                                                       & \textbf{1.42$_{0.8}$}                                             & 90.06$_{0.3}$                                                       & 2.77                                                          \\
\multicolumn{1}{l|}{\friends-G*}                      & \textbf{0.00}                                             & \textbf{91.15$_{0.4}$}                                               & 9.49$_{25.9}$                                                     & 91.06$_{0.2}$                                                       & 83.03                                                         & \textbf{1.00}                                             & 90.09$_{0.4}$                                                       & \textbf{1.37$_{0.9}$}                                              & 90.01$_{0.2}$                                                       & 3.18                                                          \\ \midrule
\multicolumn{1}{l|}{\textbf{\pebm}}                   & \textbf{0.00}                                             & \textbf{92.26$_{0.2}$}                                               & \textbf{1.27$_{0.6}$}                                              & \textbf{92.91$_{0.2}$}                                               & \textbf{2.16}                                                 & \textbf{1.00}                                             & \textbf{91.36$_{0.3}$}                                               & \textbf{1.46$_{0.8}$}                                              & \textbf{91.83$_{0.3}$}                                               & \textbf{2.49}                                                 \\
\multicolumn{1}{l|}{\textbf{\pebm-P}}                 & NA                                                          & NA                                                                & \textbf{1.38$_{0.7}$}                                              & \textbf{92.70$_{0.2}$}                                               & \textbf{2.78}                                                 & NA                                                          & NA                                                                & \textbf{1.63$_{1.0}$}                                              & \textbf{91.49$_{0.3}$}                                               & 3.47                                                          \\
\multicolumn{1}{l|}{\textbf{\pebm$_{CN-10}$}}          & \textbf{0.00}                                             & \textbf{92.99$_{0.2}$}                                               & \textbf{1.43$_{0.8}$}                                              & \textbf{92.90$_{0.2}$}                                               & \textbf{3.06}                                                 & \textbf{1.00}                                             & \textbf{92.02$_{0.2}$}                                               & \textbf{1.50$_{0.9}$}                                              & \textbf{92.03$_{0.2}$}                                               & \textbf{2.52}                                                 \\
\multicolumn{1}{l|}{\textbf{\pebm$_{IN}$}} & 1.00                                                      & \textbf{92.98$_{0.2}$}                                               & \textbf{1.39$_{0.8}$}                                              & \textbf{92.92$_{0.2}$}                                               & \textbf{2.50}                                                 & \textbf{1.00}                                             & \textbf{92.02$_{0.2}$}                                               & \textbf{1.52$_{0.8}$}                                              & \textbf{92.02$_{0.3}$}                                               & \textbf{2.81}                                                 \\
\multicolumn{1}{l|}{\textbf{\pebm-P$_{CN-10}$}}       & NA                                                          & NA                                                                & \textbf{1.64$_{0.01}$}                                             & \textbf{92.86$_{0.20}$}                                              & \textbf{4.34}                                                 & NA                                                          & NA                                                                & \textbf{1.68$_{1.0}$}                                              & \textbf{92.07$_{0.2}$}                                               & 3.34                                                          \\ \bottomrule
\end{tabular}
}
    % Second Table
    % \hspace{10pt}
    \scalebox{0.66}{% Please add the following required packages to your document preamble:
\begin{tabular}{@{}rccccccccc@{}}
\toprule
\multicolumn{10}{c}{\textbf{Transfer Learning}}                                                                                                                                                                                                                                                                                                                                                                                                                                                                                                                                                                                                                           \\ \midrule
\textbf{}                                                   & \multicolumn{5}{c}{\textbf{Fine-Tune}}                                                                                                                                                                                                                                                                                                  & \multicolumn{4}{c}{\textbf{Linear - Bullseye Polytope}}                                                                                                                                                                                                           \\ \cmidrule(lr){2-6} \cmidrule(lr){7-10}
\textbf{}                                                   & \multicolumn{2}{c}{\textbf{Bullseye Polytope-10\%}}                                                                             & \multicolumn{3}{c}{\textbf{Narcissus-10\%}}                                                                                                                                                           & \multicolumn{2}{c}{\textbf{BlackBox-10\%}}                                                                                      & \multicolumn{2}{c}{\textbf{WhiteBox-1\%}}                                                                                       \\ \cmidrule(lr){2-3} \cmidrule(lr){4-6}\cmidrule(lr){7-8} \cmidrule(lr){9-10}
\textbf{}                                                   & \begin{tabular}[c]{@{}c@{}}Poison \\ Success (\%) ↓\end{tabular} & \begin{tabular}[c]{@{}c@{}}Avg Natural \\ Accuracy (\%) ↑\end{tabular} & \begin{tabular}[c]{@{}c@{}}Avg Poison \\ Success (\%) ↓\end{tabular} & \begin{tabular}[c]{@{}c@{}}Avg Natural \\ Accuracy (\%) ↑\end{tabular} & \begin{tabular}[c]{@{}c@{}}Max Poison \\ Success (\%) ↓\end{tabular} & \begin{tabular}[c]{@{}c@{}}Poison \\ Success (\%) ↓\end{tabular} & \begin{tabular}[c]{@{}c@{}}Avg Natural \\ Accuracy (\%) ↑\end{tabular} & \begin{tabular}[c]{@{}c@{}}Poison \\ Success (\%) ↓\end{tabular} & \begin{tabular}[c]{@{}c@{}}Avg Natural \\ Accuracy (\%) ↑\end{tabular} \\ \midrule
\multicolumn{1}{l|}{None}                                   & 46.00                                                     & 89.84$_{0.9}$                                                       & 33.41$_{33.9}$                                                    & 90.14$_{2.4}$                                                       & 98.27                                                         & 93.75                                                     & 83.59$_{2.4}$                                                       & 98.00                                                     & 70.09$_{0.2}$                                                       \\
\multicolumn{1}{l|}{\epic-0.1}               & 50.00                                                     & 89.00$_{1.8}$                                                       & 32.40$_{33.7}$                                                    & 90.02$_{2.2}$                                                       & 98.95                                                         & 91.67                                                     & 83.48$_{2.9}$                                                       & 98.00                                                     & 69.35$_{0.3}$                                                       \\
\multicolumn{1}{l|}{\epic-0.2*}              & 42.00                                                     & 81.95$_{5.6}$                                                       & 20.93$_{27.1}$                                                    & 88.58$_{2.0}$                                                       & 91.72                                                         & 66.67                                                     & 84.34$_{3.8}$                                                       & 91.00                                                     & 64.79$_{0.7}$                                                       \\
\multicolumn{1}{l|}{\epic-0.3}      & 44.00                                                     & 86.75$_{6.3}$                                                       & 28.01$_{34.9}$                                                    & 84.36$_{6.3}$                                                       & 99.91                                                         & 66.67                                                     & 83.23$_{3.8}$                                                       & 63.00                                                     & 60.86$_{1.5}$                                                       \\
\multicolumn{1}{l|}{\friends-B}              & 8.00                                                      & 87.80$_{1.1}$                                                       & 3.34$_{5.7}$                                                      & 89.62$_{0.5}$                                                       & 19.48                                                         & 35.42                                                     & 84.97$_{2.2}$                                                       & 19.00                                                     & 60.85$_{0.6}$                                                       \\
\multicolumn{1}{l|}{\friends-G*}             & 8.00                                                      & 87.82$_{1.2}$                                                       & 3.04$_{5.1}$                                                      & 89.81$_{0.5}$                                                       & 17.32                                                         & 33.33                                                     & 85.18$_{2.3}$                                                       & 19.00                                                     & 60.90$_{0.6}$                                                       \\ \midrule
\multicolumn{1}{l|}{\textbf{\pebm}}          & \textbf{0.00}                                             & \textbf{88.95$_{1.1}$}                                              & \textbf{1.98$_{1.7}$}                                             & \textbf{91.40$_{0.4}$}                                              & \textbf{5.98}                                                 & \textbf{0.00}                                             & \textbf{92.89$_{0.2}$}                                              & \textbf{6.00}                                             & \textbf{64.51$_{0.6}$}                                              \\
\multicolumn{1}{l|}{\textbf{\pebm-P}}        & NA                                                          & NA                                                                & \textbf{3.66$_{4.63}$}                                                    & \textbf{90.89$_{0.31}$}                                                       & 16.04                                                         & NA                                                          & NA                                                                & NA                                                          & NA                                                                \\
\multicolumn{1}{l|}{\textbf{\pebm$_{CN-10}$}} & \textbf{0.00}                                             & \textbf{88.67$_{1.2}$}                                              & \textbf{2.97$_{2.5}$}                                             & \textbf{90.99$_{0.3}$}                                              & \textbf{7.95}                                                 & \textbf{0.00}                                             & \textbf{92.82$_{0.1}$}                                              & \textbf{6.00}                                             & \textbf{64.44$_{0.4}$}                                              \\
\multicolumn{1}{l|}{\textbf{\pebm$_{IN}$}}             & \textbf{0.00}                                             & \textbf{87.52$_{1.2}$}                                              & \textbf{2.02$_{1.0}$}                                             & \textbf{89.78$_{0.6}$}                                              & \textbf{3.85}                                                 & \textbf{0.00}                                             & \textbf{92.38$_{0.3}$}                                              & \textbf{6.00}                                             & \textbf{64.98$_{0.3}$}                                              \\ \bottomrule
\end{tabular}}
    \label{table:core_full_results}
\end{table}

\subsection{Extended Poison\% Results}\label{app:results_poison_pct}

\begin{table}[H]
   \centering
   \caption{Narcissus transfer fine-tune results at various poison\%'s}
   \label{table:narc_poison_pct}
   \scalebox{0.6}{% Please add the following required packages to your document preamble:
\begin{tabular}{@{}rccccccccc@{}}
\toprule
\multicolumn{1}{c}{Poison-\%}                        & \multicolumn{3}{c}{\textbf{1\%}}                                                                                                                                                                                             & \multicolumn{3}{c}{\textbf{2.5\%}}                                                                                                                                                                                           & \multicolumn{3}{c}{\textbf{10\%}}                                                                                                                                                                                            \\ \cmidrule(lr){2-4} \cmidrule(lr){5-7} \cmidrule(lr){8-10} 
                                                     & \textbf{\begin{tabular}[c]{@{}c@{}}Avg Poison \\ Success (\%) ↓\end{tabular}} & \textbf{\begin{tabular}[c]{@{}c@{}}Avg Natural\\ Accuracy (\%) ↑\end{tabular}} & \textbf{\begin{tabular}[c]{@{}c@{}}Max Poison\\ Success (\%) ↓\end{tabular}} & \textbf{\begin{tabular}[c]{@{}c@{}}Avg Poison \\ Success (\%) ↓\end{tabular}} & \textbf{\begin{tabular}[c]{@{}c@{}}Avg Natural\\ Accuracy (\%) ↑\end{tabular}} & \textbf{\begin{tabular}[c]{@{}c@{}}Max Poison\\ Success (\%) ↓\end{tabular}} & \textbf{\begin{tabular}[c]{@{}c@{}}Avg Poison \\ Success (\%) ↓\end{tabular}} & \textbf{\begin{tabular}[c]{@{}c@{}}Avg Natural\\ Accuracy (\%) ↑\end{tabular}} & \textbf{\begin{tabular}[c]{@{}c@{}}Max Poison\\ Success (\%) ↓\end{tabular}} \\ \midrule
\multicolumn{1}{l|}{None}                            & 17.06$_{27.0}$                                                             & 93.18$_{0.1}$                                                             & 81.97                                                                 & 22.22$_{30.1}$                                                             & 93.35$_{0.1}$                                                             & 89.74                                                                 & 33.41$_{33.9}$                                                             & 90.14$_{2.4}$                                                             & 98.27                                                                 \\
\multicolumn{1}{l|}{\epic-0.1}        & 15.58$_{25.5}$                                                             & 92.75$_{0.2}$                                                             & 73.65                                                                 & 19.77$_{27.5}$                                                             & 92.72$_{0.3}$                                                             & 87.51                                                                 & 32.40$_{33.7}$                                                             & 90.02$_{2.2}$                                                             & 98.95                                                                 \\
\multicolumn{1}{l|}{\epic-0.2}        & 12.33$_{23.8}$                                                             & 85.86$_{2.9}$                                                            & 74.32                                                                 & 24.26$_{31.2}$                                                             & 85.59$_{3.3}$                                                             & 96.07                                                                 & 20.93$_{27.1}$                                                             & 88.58$_{2.0}$                                                             & 91.72                                                                 \\
\multicolumn{1}{l|}{\epic-0.3}        & 12.74$_{21.2}$                                                             & 91.37$_{4.0}$                                                             & 67.45                                                                 & 12.32$_{18.7}$                                                             & 92.24$_{0.4}$                                                             & 61.33                                                                 & 28.01$_{34.9}$                                                             & 84.36$_{6.3}$                                                             & 99.91                                                                 \\
\multicolumn{1}{l|}{\friends-B}       & \textbf{1.44$_{0.8}$}                                                      & 90.61$_{0.2}$                                                             & \textbf{2.49}                                                         & 2.25$_{3.3}$                                                               & 90.44$_{0.3}$                                                             & 11.46                                                                 & 3.34$_{5.7}$                                                               & 89.62$_{0.5}$                                                             & 19.48                                                                 \\
\multicolumn{1}{l|}{\friends-G}       & \textbf{1.34$_{0.7}$}                                                      & 90.50$_{0.2}$                                                             & \textbf{2.50}                                                         & 2.43$_{3.6}$                                                               & 90.51$_{0.2}$                                                             & 12.61                                                                 & 3.04$_{5.1}$                                                               & 89.81$_{0.5}$                                                             & 17.32                                                                 \\ \midrule
\multicolumn{1}{l|}{\textbf{\pebm}}   & \textbf{1.50$_{1.4}$}                                                      & \textbf{91.65$_{0.1}$}                                                    & 5.19                                                                  & \textbf{1.60$_{1.2}$}                                                      & \textbf{91.27$_{0.1}$}                                                    & \textbf{4.76}                                                         & \textbf{1.98$_{1.7}$}                                                      & \textbf{91.40$_{0.4}$}                                                    & \textbf{5.98}                                                         \\
\multicolumn{1}{l|}{\textbf{\pebm-P}} & 4.50$_{7.4}$                                                               & 89.61$_{0.3}$                                                             & 24.43                                                                 & 7.93$_{12.4}$                                                              & 90.26$_{0.2}$                                                             & 39.59                                                                 & \textbf{3.66$_{4.63}$}                                                    & \textbf{90.89$_{0.31}$}                                                       & 16.04                                                                 \\
\multicolumn{1}{l|}{\textbf{\pebm$_{CN-10}$}}   & \textbf{1.77$_{1.2}$}                                                      & \textbf{91.56$_{0.1}$}                                                    & 4.07                                                                  & \textbf{2.21$_{1.6}$}                                                      & \textbf{91.45$_{0.1}$}                                                    & \textbf{5.02}                                                         & \textbf{2.97$_{2.5}$}                                                      & \textbf{90.99$_{0.3}$}                                                    & \textbf{7.95}                                                         \\
\multicolumn{1}{l|}{\textbf{\pebm$_{IN}$}}   & \textbf{1.62$_{0.9}$}                                                      & 90.91$_{0.1}$                                                             & 3.35                                                                  & \textbf{1.85$_{0.9}$}                                                      & 90.85$_{0.2}$                                                             & \textbf{3.39}                                                         & \textbf{2.02$_{1.0}$}                                                      & 89.78$_{0.6}$                                                             & \textbf{3.85}                                                         \\
\multicolumn{1}{l|}{\textbf{\pebm-P$_{CN-10}$}} & 4.33$_{6.2}$                                                               & 90.99$_{0.2}$                                                             & 21.25                                                                 & 5.95$_{8.5}$                                                               & 90.80$_{0.2}$                                                             & 28.88                                                                 & 11.84$_{19.9}$                                                             & 88.77$_{1.3}$                                                             & 66.63                                                                 \\ \bottomrule
\end{tabular}}
\end{table}

\begin{table}[ht]
   \centering
   \caption{BP transfer linear gray-box results at various poison\%'s}
   \label{table:bp_poison_pct}
   \scalebox{0.7}{% Please add the following required packages to your document preamble:

\begin{tabular}{@{}lcccccccc@{}}
\toprule
\multicolumn{1}{c}{Poison-\%}                            & \multicolumn{2}{c}{\textbf{1\%}}                                                                                                               & \multicolumn{2}{c}{\textbf{2\%}}                                                                                                              & \multicolumn{2}{c}{\textbf{5\%}}                                                                                                              & \multicolumn{2}{c}{\textbf{10\%}}                                                                                                              \\ \cmidrule(lr){2-3} \cmidrule(lr){4-5} \cmidrule(lr){6-7} \cmidrule(lr){8-9} 
                                                         & \textbf{\begin{tabular}[c]{@{}c@{}}Poison \\ Success (\%) ↓\end{tabular}} & \textbf{\begin{tabular}[c]{@{}c@{}}Natural\\ Accuracy (\%) ↑\end{tabular}} & \textbf{\begin{tabular}[c]{@{}c@{}}Poison \\ Success (\%) ↓\end{tabular}} & \textbf{\begin{tabular}[c]{@{}c@{}}Natural\\ Accuracy (\%) ↑\end{tabular}} & \textbf{\begin{tabular}[c]{@{}c@{}}Poison \\ Success (\%) ↓\end{tabular}} & \textbf{\begin{tabular}[c]{@{}c@{}}Natural\\ Accuracy (\%) ↑\end{tabular}} & \textbf{\begin{tabular}[c]{@{}c@{}}Poison \\ Success (\%) ↓\end{tabular}} & \textbf{\begin{tabular}[c]{@{}c@{}}Natural\\ Accuracy (\%) ↑\end{tabular}} \\ \midrule
\multicolumn{1}{l|}{None}                                & 26.00                                                              & 93.60$_{0.2}$                                                           & 32.00                                                              & 93.60$_{0.2}$                                                           & 66.00                                                              & 92.89$_{0.4}$                                                           & 93.75                                                              & 83.59$_{2.4}$                                                           \\
\multicolumn{1}{l|}{\epic-0.1}            & 12.00                                                              & 93.34$_{0.4}$                                                           & 50.00                                                              & 92.79$_{0.6}$                                                           & 70.00                                                              & 92.43$_{0.8}$                                                           & 91.67                                                              & 83.48$_{2.9}$                                                           \\
\multicolumn{1}{l|}{\epic-0.2}            & 18.00                                                              & 92.53$_{1.4}$                                                           & 34.00                                                              & 92.86$_{1.4}$                                                           & 76.00                                                              & 91.72$_{2.0}$                                                           & 66.67                                                              & 84.34$_{3.8}$                                                           \\
\multicolumn{1}{l|}{\epic-0.3}            & 18.00                                                              & 92.80$_{0.9}$                                                           & 24.00                                                              & 92.89$_{1.0}$                                                           & 62.00                                                              & 90.95$_{2.7}$                                                           & 66.67                                                              & 83.23$_{3.8}$                                                           \\
\multicolumn{1}{l|}{\friends-B}           & 4.00                                                               & \textbf{94.09$_{0.1}$}                                                  & 4.00                                                               & \textbf{94.11$_{0.1}$}                                                  & 26.00                                                              & \textbf{93.72$_{0.2}$}                                                  & 35.42                                                              & 84.97$_{2.2}$                                                           \\
\multicolumn{1}{l|}{\friends-G}           & 4.00                                                               & \textbf{94.12$_{0.1}$}                                                  & 4.00                                                               & \textbf{94.13$_{0.1}$}                                                  & 22.00                                                              & \textbf{93.73$_{0.2}$}                                                  & 33.33                                                              & 85.18$_{2.3}$                                                           \\ \midrule
\multicolumn{1}{l|}{\textbf{\pebm}}       & \textbf{0.00}                                                      & 93.18$_{0.0}$                                                           & \textbf{0.00}                                                      & 92.94$_{0.1}$                                                           & \textbf{0.00}                                                      & 92.92$_{0.1}$                                                           & \textbf{0.00}                                                      & \textbf{92.89$_{0.2}$}                                                  \\
\multicolumn{1}{l|}{\textbf{\pebm$_{CN-10}$}} & \textbf{0.00}                                                      & 93.14$_{0.1}$                                                           & \textbf{0.00}                                                      & 92.61$_{0.1}$                                                           & \textbf{0.00}                                                      & 93.00$_{0.1}$                                                           & \textbf{0.00}                                                      & \textbf{92.82$_{0.1}$}                                                  \\
\multicolumn{1}{l|}{\textbf{\pebm$_{IN}$}}                   & \textbf{0.00}                                                      & 92.09$_{0.1}$                                                           & \textbf{0.00}                                                      & 91.51$_{0.1}$                                                           & \textbf{0.00}                                                      & 92.75$_{0.1}$                                                           & \textbf{0.00}                                                      & \textbf{92.38$_{0.3}$}                                                  \\ \bottomrule
\end{tabular}
}
\end{table}

\subsection{Full MobileNetV2 and DenseNet121 Results}\label{app:models_full}

\begin{table}[H]
    \centering
    \caption{MobileNetV2 Full Results}
    % First Table
    \scalebox{0.6}{% Please add the following required packages to your document preamble:
\begin{tabular}{rcccccccccc}
\toprule
\multicolumn{11}{c}{\textbf{From Scratch - MobileNetV2}}                                                                                                                                                                                                                                                                                                                                                                                                                                                                                                                                                                                                                                                                                                                                                                                                                                                                                       \\ \midrule
\textbf{}                                          & \multicolumn{5}{c}{\textbf{200 - Epochs}}                                                                                                                                                                                                                                                                                                                                                                                                   & \multicolumn{5}{c}{\textbf{80 - Epochs}}                                                                                                                                                                                                                                                                                                                                                                                                    \\ \cmidrule(lr){2-6} \cmidrule(lr){7-11}
\textbf{}                                          & \multicolumn{2}{c}{\textbf{Gradient Matching-1\%}}                                                                                                                      & \multicolumn{3}{c}{\textbf{Narcissus-1\%}}                                                                                                                                                                                                                        & \multicolumn{2}{c}{\textbf{Gradient Matching-1\%}}                                                                                                                      & \multicolumn{3}{c}{\textbf{Narcissus-1\%}}                                                                                                                                                                                                                        \\ \cmidrule(lr){2-3} \cmidrule(lr){4-6}\cmidrule(lr){7-8}\cmidrule(lr){9-11}
\textbf{}                                          & \multicolumn{1}{c}{\begin{tabular}[c]{@{}c@{}}Poison \\ Success (\%) ↓\end{tabular}} & \multicolumn{1}{c}{\begin{tabular}[c]{@{}c@{}}Avg Natural \\ Accuracy (\%) ↑\end{tabular}} & \multicolumn{1}{c}{\begin{tabular}[c]{@{}c@{}}Avg Poison \\ Success (\%) ↓\end{tabular}} & \multicolumn{1}{c}{\begin{tabular}[c]{@{}c@{}}Avg Natural \\ Accuracy (\%) ↑\end{tabular}} & \multicolumn{1}{c}{\begin{tabular}[c]{@{}c@{}}Max Poison \\ Success (\%) ↓\end{tabular}} & \multicolumn{1}{c}{\begin{tabular}[c]{@{}c@{}}Poison \\ Success (\%) ↓\end{tabular}} & \multicolumn{1}{c}{\begin{tabular}[c]{@{}c@{}}Avg Natural \\ Accuracy (\%) ↑\end{tabular}} & \multicolumn{1}{c}{\begin{tabular}[c]{@{}c@{}}Avg Poison \\ Success (\%) ↓\end{tabular}} & \multicolumn{1}{c}{\begin{tabular}[c]{@{}c@{}}Avg Natural \\ Accuracy (\%) ↑\end{tabular}} & \multicolumn{1}{c}{\begin{tabular}[c]{@{}c@{}}Max Poison \\ Success (\%) ↓\end{tabular}} \\ \midrule
\multicolumn{1}{r|}{None}                          & 20.00                                                                         & 93.86$_{0.2}$                                                                           & 32.70$_{24.5}$                                                                        & 93.92$_{0.1}$                                                                           & 73.97                                                                             & 30.00                                                                         & 92.54$_{0.2}$                                                                           & 27.26$_{26.5}$                                                                        & 92.53$_{0.2}$                                                                           & 74.82                                                                             \\
\multicolumn{1}{r|}{\epic-0.1}      & 37.50                                                                         & 91.28$_{0.2}$                                                                           & 40.09$_{27.1}$                                                                        & 91.15$_{0.2}$                                                                           & 79.74                                                                             & 16.00                                                                         & 90.45$_{0.3}$                                                                           & 31.37$_{30.9}$                                                                        & 90.51$_{0.3}$                                                                           & 89.36                                                                             \\
\multicolumn{1}{r|}{\epic-0.2}      & 19.00                                                                         & 91.24$_{0.2}$                                                                           & 38.55$_{27.5}$                                                                        & 87.65$_{0.5}$                                                                           & 74.72                                                                             & 22.00                                                                         & 89.90$_{0.3}$                                                                           & 29.22$_{27.6}$                                                                        & 89.91$_{0.3}$                                                                           & 76.54                                                                             \\
\multicolumn{1}{r|}{\epic-0.3}      & 9.78                                                                          & 87.80$_{1.6}$                                                                           & 22.35$_{23.9}$                                                                        & 78.16$_{9.9}$                                                                           & 69.52                                                                             & 14.00                                                                         & 90.23$_{0.3}$                                                                           & 30.69$_{30.6}$                                                                        & 90.30$_{0.3}$                                                                           & 82.92                                                                             \\
\multicolumn{1}{r|}{\friends-B}     & 6.00                                                                          & 84.30$_{2.7}$                                                                           & \textbf{2.00$_{1.3}$}                                                                 & 88.82$_{0.6}$                                                                           & 4.88                                                                              & \textbf{1.00}                                                                 & 87.89$_{0.3}$                                                                           & \textbf{1.98$_{1.1}$}                                                                 & 87.90$_{0.4}$                                                                           & 4.00                                                                              \\
\multicolumn{1}{r|}{\friends-G}     & 5.00                                                                          & 88.84$_{0.4}$                                                                           & \textbf{2.05$_{1.7}$}                                                                 & 88.93$_{0.3}$                                                                           & 6.33                                                                              & 3.00                                                                          & 87.90$_{0.4}$                                                                           & \textbf{2.00$_{1.4}$}                                                                 & 88.09$_{0.3}$                                                                           & 5.07                                                                              \\
\multicolumn{1}{r|}{\textbf{\pebm}} & \textbf{1.00}                                                                 & \textbf{90.93$_{0.2}$}                                                                  & \textbf{1.64$_{0.8}$}                                                                 & \textbf{91.75$_{0.1}$}                                                                  & \textbf{2.91}                                                                     & \textbf{1.00}                                                                 & \textbf{89.71$_{0.2}$}                                                                  & \textbf{1.79$_{0.8}$}                                                                 & \textbf{90.64$_{0.2}$}                                                                  & \textbf{2.65}                                                                     \\ \bottomrule
\end{tabular}
}
    % Second Table
    \scalebox{0.66}{% Please add the following required packages to your document preamble:
\begin{tabular}{lccccc}
\toprule
\multicolumn{6}{c}{\textbf{Transfer Learning - MobileNetV2}}                                                                                                                                                                                                                                                                                                                                                                                                                                         \\ \midrule
\textbf{}                                              & \multicolumn{3}{c}{\textbf{Fine-Tune NS-10\%}}                                                                                                                                                                                                                            & \multicolumn{2}{c}{\textbf{Transfer Linear BP BlackBox-10\%}}                                                                                                                 \\ \cmidrule(lr){2-4} \cmidrule(lr){5-6} 
\textbf{}                                              & \multicolumn{1}{c}{\begin{tabular}[c]{@{}c@{}}Avg Poison \\ Success (\%) ↓\end{tabular}} & \multicolumn{1}{c}{\begin{tabular}[c]{@{}c@{}}Avg Natural \\ Accuracy (\%) ↑\end{tabular}} & \multicolumn{1}{c}{\begin{tabular}[c]{@{}c@{}}Max Poison \\ Success (\%) ↓\end{tabular}} & \multicolumn{1}{c}{\begin{tabular}[c]{@{}c@{}}Poison \\ Success (\%) ↓\end{tabular}} & \multicolumn{1}{c}{\begin{tabular}[c]{@{}c@{}}Avg Natural \\ Accuracy (\%) ↑\end{tabular}} \\ \midrule
\multicolumn{1}{l|}{None}                              & 23.59$_{23.2}$                                                                        & 88.30$_{1.2}$                                                                           & 66.54                                                                             & 81.25                                                                         & 73.27$_{1.0}$                                                                           \\
\multicolumn{1}{l|}{\epic-0.1}          & 23.25$_{22.8}$                                                                        & 88.35$_{1.0}$                                                                           & 65.97                                                                             & 81.25                                                                         & 69.78$_{2.0}$                                                                           \\
\multicolumn{1}{l|}{\epic-0.2}          & 19.95$_{19.2}$                                                                        & 87.67$_{1.3}$                                                                           & 50.05                                                                             & 56.25                                                                         & 54.47$_{5.6}$                                                                           \\
\multicolumn{1}{l|}{\epic-0.3} & 21.70$_{28.1}$                                                                        & 78.17$_{6.0}$                                                                           & 74.96                                                                             & 58.33                                                                         & 58.74$_{9.0}$                                                                           \\
\multicolumn{1}{l|}{\friends-B}         & \textbf{2.21$_{1.5}$}                                                                 & \textbf{83.05$_{0.7}$}                                                                  & \textbf{5.63}                                                                     & 41.67                                                                         & 68.86$_{1.5}$                                                                           \\
\multicolumn{1}{l|}{\friends-G}         & \textbf{2.20$_{1.4}$}                                                                 & \textbf{83.04$_{0.7}$}                                                                  & \textbf{5.42}                                                                     & 47.92                                                                         & 68.94$_{1.5}$                                                                           \\
\multicolumn{1}{l|}{\textbf{\pebm}}     & \textbf{3.66$_{5.4}$}                                                                 & \textbf{84.18$_{0.5}$}                                                                  & 18.85                                                                             & \textbf{0.00}                                                                 & \textbf{78.57$_{1.4}$}                                                                  \\ \bottomrule
\end{tabular}}
    \label{table:mobile_full_results}
\end{table}

\begin{table}[H]
    \centering
    \caption{DenseNet121 Full Results}
    % First Table
    \scalebox{0.55}{% Please add the following required packages to your document preamble:
\begin{tabular}{@{}lcccccccccc@{}}
\toprule
\multicolumn{11}{c}{\textbf{From Scratch - DenseNet121}}                                                                                                                                                                                                                                                                                                                                                                                                                                                                                                                                                                                                                                                                                                                           \\ \midrule
\textbf{}                                      & \multicolumn{5}{c}{\textbf{200 - Epochs}}                                                                                                                                                                                                                                                                                                                       & \multicolumn{5}{c}{\textbf{80 - Epochs}}                                                                                                                                                                                                                                                                                                                        \\ \cmidrule(l){2-11} 
\textbf{}                                      & \multicolumn{2}{c}{\textbf{Gradient Matching-1\%}}                                                                                        & \multicolumn{3}{c}{\textbf{Narcissus-1\%}}                                                                                                                                                                          & \multicolumn{2}{c}{\textbf{Gradient Matching-1\%}}                                                                                        & \multicolumn{3}{c}{\textbf{Narcissus-1\%}}                                                                                                                                                                          \\ \cmidrule(l){2-11} 
\textbf{}                                      & \begin{tabular}[c]{@{}c@{}}Poison \\ Success (\%) ↓\end{tabular} & \begin{tabular}[c]{@{}c@{}}Avg Natural \\ Accuracy (\%) ↑\end{tabular} & \begin{tabular}[c]{@{}c@{}}Avg Poison \\ Success (\%) ↓\end{tabular} & \begin{tabular}[c]{@{}c@{}}Avg Natural \\ Accuracy (\%) ↑\end{tabular} & \begin{tabular}[c]{@{}c@{}}Max Poison \\ Success (\%) ↓\end{tabular} & \begin{tabular}[c]{@{}c@{}}Poison \\ Success (\%) ↓\end{tabular} & \begin{tabular}[c]{@{}c@{}}Avg Natural \\ Accuracy (\%) ↑\end{tabular} & \begin{tabular}[c]{@{}c@{}}Avg Poison \\ Success (\%) ↓\end{tabular} & \begin{tabular}[c]{@{}c@{}}Avg Natural \\ Accuracy (\%) ↑\end{tabular} & \begin{tabular}[c]{@{}c@{}}Max Poison \\ Success (\%) ↓\end{tabular} \\ \midrule
\multicolumn{1}{l|}{None}                      & 14.00                                                            & 95.30$_{0.1}$                                                          & $46.52_{32.2}$                                                      & 95.33$_{0.1}$                                                          & 91.96                                                               & 19.00                                                            & 94.38$_{0.2}$                                                          & 38.01$_{36.3}$                                                       & 94.49$_{0.1}$                                                          & 89.11                                                               \\
\multicolumn{1}{l|}{\epic-0.1}  & 14.00                                                            & 93.0$_{0.3}$                                                           & $43.38_{32.0}$                                                           & 93.07$_{0.2}$                                                          & 88.97                                                               & 16.00                                                            & 92.78$_{0.3}$                                                          & 32.85$_{33.0}$                                                       & 92.87$_{0.3}$                                                          & 79.42                                                               \\
\multicolumn{1}{l|}{\epic-0.2}  & 7.00                                                             & 90.67$_{0.5}$                                                          & $41.97_{33.2}$                                                           & 90.23$_{0.6}$                                                          & 86.85                                                               & 13.00                                                            & 92.69$_{0.3}$                                                          & 30.67$_{28.1}$                                                       & 92.82$_{0.2}$                                                          & 65.46                                                               \\
\multicolumn{1}{l|}{\epic-0.3}  & 4.00                                                             & 88.3$_{1.0}$                                                           & $32.60_{29.4}$                                                           & 85.12$_{2.4}$                                                          & 71.50                                                               & 15.00                                                            & 93.35$_{0.2}$                                                          & 36.80$_{36.0}$                                                       & 93.34$_{0.2}$                                                          & 90.41                                                               \\
\multicolumn{1}{l|}{\friends-B} & 1.00                                                             & \textbf{91.33$_{0.4}$}                                                 & $8.60_{21.2}$                                                            & 91.55$_{0.3}$                                                          & 68.57                                                               & \textbf{1.00}                                                    & 89.93$_{0.4}$                                                          & 5.60$_{11.6}$                                                        & 90.01$_{0.4}$                                                          & 38.08                                                               \\
\multicolumn{1}{l|}{\friends-G} & 1.00                                                             & \textbf{91.33$_{0.4}$}                                                 & $10.13_{25.2}$                                                           & 91.32$_{0.4}$                                                          & 81.47                                                               & \textbf{1.00}                                                    & 89.97$_{0.4}$                                                          & 7.59$_{18.7}$                                                        & 89.89$_{0.4}$                                                          & 60.68                                                               \\
\multicolumn{1}{l|}{\textbf{\pebm}}                  & \textbf{0.00}                                                    & \textbf{92.85$_{0.2}$}                                                 & \textbf{$1.42_{0.7}$}                                                    & \textbf{93.48$_{0.1}$}                                                 & \textbf{2.60}                                                       & 2.00                                                             & \textbf{91.88$_{0.3}$}                                                 & \textbf{1.59$_{0.9}$}                                                & \textbf{92.59$_{0.2}$}                                                 & \textbf{3.06}                                                       \\ \bottomrule
\end{tabular}}
    % Second Table
    % \hspace{10pt}
    \scalebox{0.7}{% Please add the following required packages to your document preamble:
\begin{tabular}{@{}lccccccc@{}}
\toprule
\multicolumn{8}{c}{\textbf{Transfer Learning - DenseNet121}}                                                                                                                                                                                                                                                                                                                                                                                                                                                                       \\ \midrule
\textbf{}                                              & \multicolumn{5}{c}{\textbf{Fine-Tune}}                                                                                                                                                                                                                                                                                                  & \multicolumn{2}{c}{\textbf{Linear}}                                                                         \\ \cmidrule(lr){2-6}\cmidrule(lr){7-8} 
\textbf{}                                              & \multicolumn{2}{c}{\textbf{Bullseye Polytope-10\%}}                                                                             & \multicolumn{3}{c}{\textbf{Narcissus-10\%}}                                                                                                                                                           & \multicolumn{2}{c}{\textbf{Bullseye Polytope-10\%}}                                                                                      \\ \cmidrule(l){2-3}\cmidrule(lr){4-6}\cmidrule(lr){7-8} 
\textbf{}                                              & \begin{tabular}[c]{@{}c@{}}Poison \\ Success (\%) ↓\end{tabular} & \begin{tabular}[c]{@{}c@{}}Avg Natural \\ Accuracy (\%) ↑\end{tabular} & \begin{tabular}[c]{@{}c@{}}Avg Poison \\ Success (\%) ↓\end{tabular} & \begin{tabular}[c]{@{}c@{}}Avg Natural \\ Accuracy (\%) ↑\end{tabular} & \begin{tabular}[c]{@{}c@{}}Max Poison \\ Success (\%) ↓\end{tabular} & \begin{tabular}[c]{@{}c@{}}Poison \\ Success (\%) ↓\end{tabular} & \begin{tabular}[c]{@{}c@{}}Avg Natural \\ Accuracy (\%) ↑\end{tabular} \\ \midrule
\multicolumn{1}{l|}{None}                              & 16.00                                                     & 88.91$_{0.7}$                                                       & 56.52$_{38.6}$                                                    & 87.03$_{2.8}$                                                       & 99.56                                                         & 73.47                                                     & 82.13$_{1.6}$                                                       \\
\multicolumn{1}{l|}{\epic-0.1}          & 18.00                                                     & 88.09$_{1.0}$                                                       & 53.97$_{39.0}$                                                    & 87.04$_{2.8}$                                                       & 99.44                                                         & 62.50                                                     & 78.88$_{2.1}$                                                       \\
\multicolumn{1}{l|}{\epic-0.2}          & 14.00                                                     & 80.44$_{3.1}$                                                       & 43.66$_{36.5}$                                                    & 85.97$_{2.6}$                                                       & 97.17                                                         & 41.67                                                     & 70.13$_{5.2}$                                                       \\
\multicolumn{1}{l|}{\epic-0.3} & 10.00                                                     & 72.84$_{11.9}$                                                      & 43.24$_{43.0}$                                                    & 72.76$_{10.8}$                                                      & 100.00                                                        & 66.67                                                     & 70.20$_{10.1}$                                                      \\
\multicolumn{1}{l|}{\friends-B}         & 4.00                                                      & \textbf{87.06$_{1.0}$}                                              & 5.34$_{9.9}$                                                      & \textbf{88.62$_{0.8}$}                                              & 33.42                                                         & 60.42                                                     & 80.22$_{1.9}$                                                       \\
\multicolumn{1}{l|}{\friends-G}         & 2.00                                                      & \textbf{87.37$_{0.9}$}                                              & 5.55$_{10.4}$                                                     & \textbf{88.75$_{0.6}$}                                              & 34.91                                                         & 56.25                                                     & 80.12$_{1.8}$                                                       \\
\multicolumn{1}{l|}{\textbf{\pebm}}     & \textbf{0.00}                                             & 84.39$_{1.0}$                                                       & \textbf{2.48$_{1.9}$}                                             & \textbf{88.75$_{0.5}$}                                              & \textbf{7.41}                                                 & \textbf{0.00}                                             & \textbf{89.29$_{0.9}$}                                              \\ \bottomrule
\end{tabular}
}
    \label{table:dense_full_results}
\end{table}

\subsection{Full CINIC-10 Results}\label{app:cinic_full_results}

\begin{table}[H]
    \centering
    \caption{CINIC-10 Full Results}
    % First Table
    \scalebox{0.8}{% Please add the following required packages to your document preamble:
% \usepackage{booktabs}
% \usepackage{booktabs}
\begin{tabular}{@{}lcccc@{}}
\toprule
\multicolumn{5}{c}{\textbf{CINIC-10 Narcissus - 1 From-Scratch}}                                                                                                                                                                                                                                                                                                                           \\ \midrule
                        & \multicolumn{4}{c}{\textbf{200 - Epochs}}                                                                                                                                                                                                                                                                                                    \\ \cmidrule(lr){2-5} 
                                             & \multicolumn{1}{c}{\begin{tabular}[c]{@{}c@{}}Avg Poison \\ Success (\%) ↓\end{tabular}} & \multicolumn{1}{c}{\begin{tabular}[c]{@{}c@{}}Avg Natural\\  Accuracy (\%) ↑\end{tabular}} & \multicolumn{1}{c}{\begin{tabular}[c]{@{}c@{}}Max Poison \\ Success (\%) ↓\end{tabular}} & \multicolumn{1}{c}{\begin{tabular}[c]{@{}c@{}}CIFAR-10\\ Accuracy (\%) ↑\end{tabular}} \\ \midrule
\multicolumn{1}{l|}{None}                    & 62.06$_{0.21}$                                                                      & 86.32$_{0.10}$                                                                          & 90.79                                                                           & 94.22$_{0.16}$                                                                      \\
\multicolumn{1}{l|}{\epic}    & 49.50$_{0.27}$                                                                        & 81.91$_{0.08}$                                                                          & 91.35                                                                           & 91.10$_{0.21}$                                                                      \\
\multicolumn{1}{l|}{\friends} & 11.17$_{0.25}$                                                                        & 77.53$_{0.60}$                                                                          & 82.21                                                                           & 88.27$_{0.68}$                                                                      \\
\multicolumn{1}{l|}{\pebm}    & \textbf{7.73$_{0.08}$}                                                                & \textbf{82.37$_{0.14}$}                                                               & \textbf{29.48}                                                                  & \textbf{91.98$_{0.16}$}                                                           \\ \midrule
                                             & \multicolumn{4}{c}{\textbf{80 - Epochs}}                                                                                                                                                                                                                                                                                                      \\ \cmidrule(lr){2-5} 
                                             & \multicolumn{1}{c}{\begin{tabular}[c]{@{}c@{}}Avg Poison \\ Success (\%) ↓\end{tabular}} & \multicolumn{1}{c}{\begin{tabular}[c]{@{}c@{}}Avg Natural\\  Accuracy (\%) ↑\end{tabular}} & \multicolumn{1}{c}{\begin{tabular}[c]{@{}c@{}}Max Poison \\ Success (\%) ↓\end{tabular}} & \multicolumn{1}{c}{\begin{tabular}[c]{@{}c@{}}CIFAR-10\\ Accuracy (\%) ↑\end{tabular}} \\ \midrule
\multicolumn{1}{l|}{None}                    & 43.75$_{0.25}$                                                     & 85.25$_{0.16}$                                                      & 82.63                                                         & 93.36$_{0.20}$                                                   \\
\multicolumn{1}{l|}{\epic}    & 37.35$_{0.26}$                                                  & 81.15$_{0.17}$                                                    & 79.98                                                       & 90.50$_{0.31}$                                                   \\
\multicolumn{1}{l|}{\friends} & 10.14$_{0.22}$                                                  & 77.46$_{0.54}$                                                    & 73.16                                                       & 87.79$_{0.47}$                                                \\
\multicolumn{1}{l|}{\pebm}    & \textbf{4.85$_{0.02}$}                                          & \textbf{81.65$_{0.15}$}                                           & \textbf{9.14}                                               & \textbf{91.33$_{0.20}$}                                       \\ \bottomrule
\vspace{-5mm}
\end{tabular}

}
    \label{table:cinic_full_results}
\end{table}

\newpage
\section{Further Experimental Details}

\subsection{EBM Training}

\begin{algorithm}[ht]
\caption{ML with SGD for Convergent Learning of EBM (\ref{eq:ebm})}
\begin{algorithmic}
\REQUIRE ConvNet potential $\mathcal{G}_{\theta}(x)$, number of training steps $J=150000$, initial weight $\theta_1$, training images $\{x^+_i \}_{i=1}^{N_\textrm{data}}$, data perturbation $\tau_\text{data}=0.02$, step size $\tau=0.01$, Langevin steps $T=100$, SGD learning rate $\gamma_\text{SGD}=0.00005$.
\ENSURE Weights $\theta_{J+1}$ for energy $\mathcal{G}_{\theta}(x )$.\\
\STATE
\STATE Set optimizer $g \leftarrow \text{SGD}(\gamma_\text{SGD})$. Initialize persistent image bank as $N_\text{data}$ uniform noise images.
\FOR{$j$=1:($J$+1)}
% \begin{enumerate}
    \STATE 1. Draw batch images $\{x_{(i)}^{+}\}_{i=1}^m$ from training set, where $(i)$  indicates a randomly selected index for sample $i$, and get samples $X_i^+ = x_{(i)} + \tau_\text{data} \epsilon_i$, where i.i.d. $\epsilon_{i} \sim \textrm{N}(0, I_D)$. \\
    \STATE 2. Draw initial negative samples $\{ Y_i^{(0)} \}_{i=1}^m$ from persistent image bank.
    			Update $\{ Y_i^{(0)} \}_{i=1}^m$ with the Langevin equation 
    			\begin{align*}
    			Y^{(k)}_i = Y^{(k-1)}_i -  \Delta \tau \nabla_{Y_{\tau}} f_{\theta_j}(Y_i^{\tau-1})  %\\
       % U(Y^{(k-1)}_i ; \theta_j) 
       + \sqrt{2\Delta\tau} \epsilon_{i, k} ,
                % x_{\tau+1} = x_{\tau} + \Delta \tau \nabla_{x_{\tau}} f_{\theta}(x_{\tau}) + \sqrt{2\Delta \tau} \epsilon_{\tau}  
    			\end{align*}
    			 where $\epsilon_{i , k} \sim \textrm{N}(0, I_D)$ i.i.d., for $K$ steps to obtain samples $\{ X_i^- \}_{i=1}^m = \{ Y_i^{(K)} \}_{i=1}^m$. Update persistent image bank with images $\{ Y_i^{(K)} \}_{i=1}^m$. \\
    \STATE 3. Update the weights by $\theta_{j+1} = \theta_{j} - g(\Delta \theta_j)$, where $g$ is the optimizer and 
    			\[
    			\Delta \theta_j= \frac{\partial}{\partial \theta} \left( \frac{1}{n} \sum_{i=1}^n f_{\theta_j}(X^+_i ) - \frac{1}{m}\sum_{i=1}^m f_{\theta_j}(X_i^-) \right)
    			\]
    			is the ML gradient approximation.
% \end{enumerate}
\ENDFOR
\end{algorithmic}
\label{al:ml_learning}
\end{algorithm}

Algorithm \ref{al:ml_learning} is pseudo-code for the training procedure of a data-initialized convergent EBM. We use the generator architecture of the SNGAN \cite{miyato2018spectral} for our EBM as our network architecture. \label{app:ebm_train}

\subsection{Poison Sourcing and Implementation} \label{app:poison_implement}

Triggerless attacks GM and BP poison success refers to the number of single-image targets successfully flipped to a target class (with 50 or 100 target image scenarios) while the natural accuracy is averaged across all target image training runs. Triggered attack Narcissus poison success is measured as the number of non-class samples from the test dataset shifted to the trigger class when the trigger is applied, averaged across all 10 classes, while the natural accuracy is averaged across the 10 classes on the un-triggered test data. We include the worst-defended class poison success. The Poison Success Rate for a single experiment can be defined for triggerless $PSR_{notr}$ and triggered $PSR_{tr}$ poisons as: 

\begin{equation}
    PSR_{notr}(F, i) =  \mathbbm{1}_{F(x_i^\pi) = y_i^{\text{adv}}}
\end{equation} 
\begin{equation}
    PSR_{tr}(F, y^\pi) = \frac{ \sum_{(x,y) \in \mathcal{D}_{test} \setminus \mathcal{D}_{test}^{\pi}} \mathbbm{1}_{F(x + \rho^\pi) = y^\pi}}{ \left\vert \mathcal{D}_{test} \setminus \mathcal{D}_{test}^{\pi} \right\vert }
\end{equation}

\subsubsection{Bullseye Polytope}
The Bullseye Polytope (BP) poisons are sourced from two distinct sets of authors. From the original authors of BP \cite{aghakhani2021bullseye}, we obtain poisons crafted specifically for a black-box scenario targeting ResNet18 and DenseNet121 architectures, and grey-box scenario for MobileNet (used in poison crafting). These poisons vary in the percentage of data poisoned, spanning 1\%, 2\%, 5\% and 10\%  for the linear-transfer mode and a single 1\% fine-tune mode for all models over a 500 image transfer dataset. Each of these scenarios has 50 datasets that specify a single target sample in the test-data. We also use a benchmark paper that provides a pre-trained white-box scenario on CIFAR-100 \cite{poison_baseline}. This dataset includes 100 target samples with strong poison success, but the undefended natural accuracy baseline is much lower. 

\subsubsection{Gradient Matching}
For GM, we use 100 publicly available datasets provided by \cite{geiping2021witches}. Each dataset specifies a single target image corresponding to 500 poisoned images in a target class. The goal of GM is for the poisons to move the target image into the target class, without changing too much of the remaining test dataset using gradient alignment. Therefore, each individual dataset training gives us a single datapoint of whether the target was correctly moved into the poisoned target class and the attack success rate is across all 100 datasets provided. 

\subsubsection{Narcissus}
For Narcissus triggered attack, we use the same generating process as described in the Narcissus paper, we apply the poison with a slight change to more closely match with the baseline provided by \cite{poison_baseline}. We learn a patch with $\varepsilon = 8/255$ on the entire 32-by-32 size of the image, per class, using the Narcissus generation method. We keep the number of poisoned samples comparable to GM for from-scratch experiment, where we apply the patch to 500 images (1\% of the dataset) and test on the patched dataset without the multiplier. In the fine-tune scenarios, we vary the poison\% over 1\%, 2.5\%, and 10\%, by modifying either the number of poisoned images or the transfer dataset size (specifically 20/2000, 50/2000, 50/500 poison/train samples). 

% \newpage
\subsection{Training Parameters}

We follow the training hyperparameters given by \cite{yang2022poisons,zeng2022narcissus,aghakhani2021bullseye,poison_baseline} for GM, NS, BP Black/Gray-Box, and BP White-Box respectively as closely as we can, with moderate modifications to align poison scenarios. HyperlightBench training followed the original creators settings and we only substituted in a poisoned dataloader \cite{Balsam2023hlbCIFAR10}.

\begin{table}[ht]
   \centering
   % \footnotesize
   \label{table:train_params}
   \scalebox{1}{\begin{tabular}{@{}lcccc@{}}
\toprule
\multicolumn{1}{c}{\textbf{Parameter}}                                               & \textbf{Shared} & \textbf{From Scratch}                                                                  & \textbf{Transfer Linear} & \textbf{Transfer Fine-Tune} \\ \midrule
Device Type                                                                          & TPU-V3          & -                                                                                      & -                        & -                           \\
Weight Decay                                                                         & 5e-4            & -                                                                                      & -                        & -                           \\
Batch Size                                                                           & -               & 128                                                                                    & 64                       & 128                         \\
Augmentations                                                                        & -               & RandomCrop(32, padding=4)                                                               & None                     & None                        \\
Epochs                                                                               & -               & 200 or 80                                                                              & 40                       & 60                          \\
Optimizer                                                                            & -               & SGD(momentum=0.9)                                                                     & SGD                      & Adam                        \\
Learning Rate                                                                        & -               & 0.1                                                                                    & 0.1                   &    0.0001                         \\
\begin{tabular}[c]{@{}l@{}}Learning Rate Schedule \\ (Multi-Step Decay)\end{tabular} & -               & \begin{tabular}[c]{@{}c@{}}100, 150 - 200 epochs\\ 30, 50, 70 - 80 epochs\end{tabular} & 15, 25, 35               & 15, 30, 45                  \\
Langevin Steps (EBM)                                                                 & -               & 150                                                                                    & 500                      & 1000                        \\
Langevin Temperature (EBM)                                                           & -               & $1 \times 10^{-4}$                                                                                   & $7.5\times 10^{-5}$                  & $1 \times 10^{-4}$                      \\
Reinitialize Linear Layer                                                            & -               & NA                                                                                     & True                     & True                        \\ \bottomrule
\end{tabular}}
\end{table}

\section{Timing Analysis} \label{app:timing}
Table \ref{table:train_times} shows the training times for each poison defense in the from-scratch scenario on a TPU-V3. As \pebm is a preprocessing step, the purification time ($\sim$400 seconds) is shared across poison scenarios, making it increasingly comparable to no defense as the number of models/scenarios increase. Although EBM training is a compute intensive process, noted in detail in App. \ref{app:ebm_train}, we share results in the section Table \ref{table:core_results} on how a single EBM on a POOD dataset can obtain SoTA performance in a poison/classifier agnostic way. While subset selection methods like \epic can reduce training time in longer scenarios, \pebm offers superior performance and flexibility to the classifier training pipeline. 

\begin{table}[H]
   \centering
   \footnotesize
   \caption{Median Wall Clock Train Times From Scratch}
   \label{table:train_times}
   \scalebox{0.7}{% Please add the following required packages to your document preamble:
\begin{tabular}{@{}lcccr@{}}
\toprule
\multicolumn{5}{c}{\textbf{Train Time (seconds)}}                                                                                                                                    \\ \midrule
\multicolumn{1}{c}{\textbf{}}                & \multicolumn{2}{c}{\textbf{Gradient Matching}}              & \multicolumn{2}{c}{\textbf{Narcissus}}                                  \\ \cmidrule(l){2-3} \cmidrule(l){4-5} 
\multicolumn{1}{c}{\textbf{epochs}}          & \textbf{80}                          & \textbf{200}         & \textbf{80}                          & \multicolumn{1}{c}{\textbf{200}} \\ \midrule
\multicolumn{1}{l|}{None}                    & \textbf{2202$_{16}$}                    & 5482$_{49}$              & \textbf{2936$_{94}$}                     & 7154$_{194}$                         \\
\multicolumn{1}{l|}{\epic}    & \textbf{2256$_{97}$}                     & \textbf{5006$_{253}$}    & 3564$_{213}$                             & \textbf{6359$_{462}$}                \\
\multicolumn{1}{l|}{\friends} & 7740$_{394}$         & 11254$_{413}$            & 8728$_{660}$         & 12868$_{573}$                        \\
\multicolumn{1}{l|}{\pebm}    & \textbf{2213$_{36}$} & 5520$_{47}$              & \textbf{2962$_{92}$} & 7293$_{219}$                         \\ \midrule
                                             % & \multicolumn{1}{l}{}                 & \multicolumn{1}{l}{} & \multicolumn{1}{l}{}                 & \multicolumn{1}{l}{}             \\ \bottomrule

\vspace{-10mm}
\end{tabular}
}
\end{table}

\section{Additional Model Interpretability Results}\label{app:captum}

\begin{figure}[H]
    \centering
    \begin{subfigure}{0.7\linewidth}
        \centering
        \includegraphics[width=\linewidth]{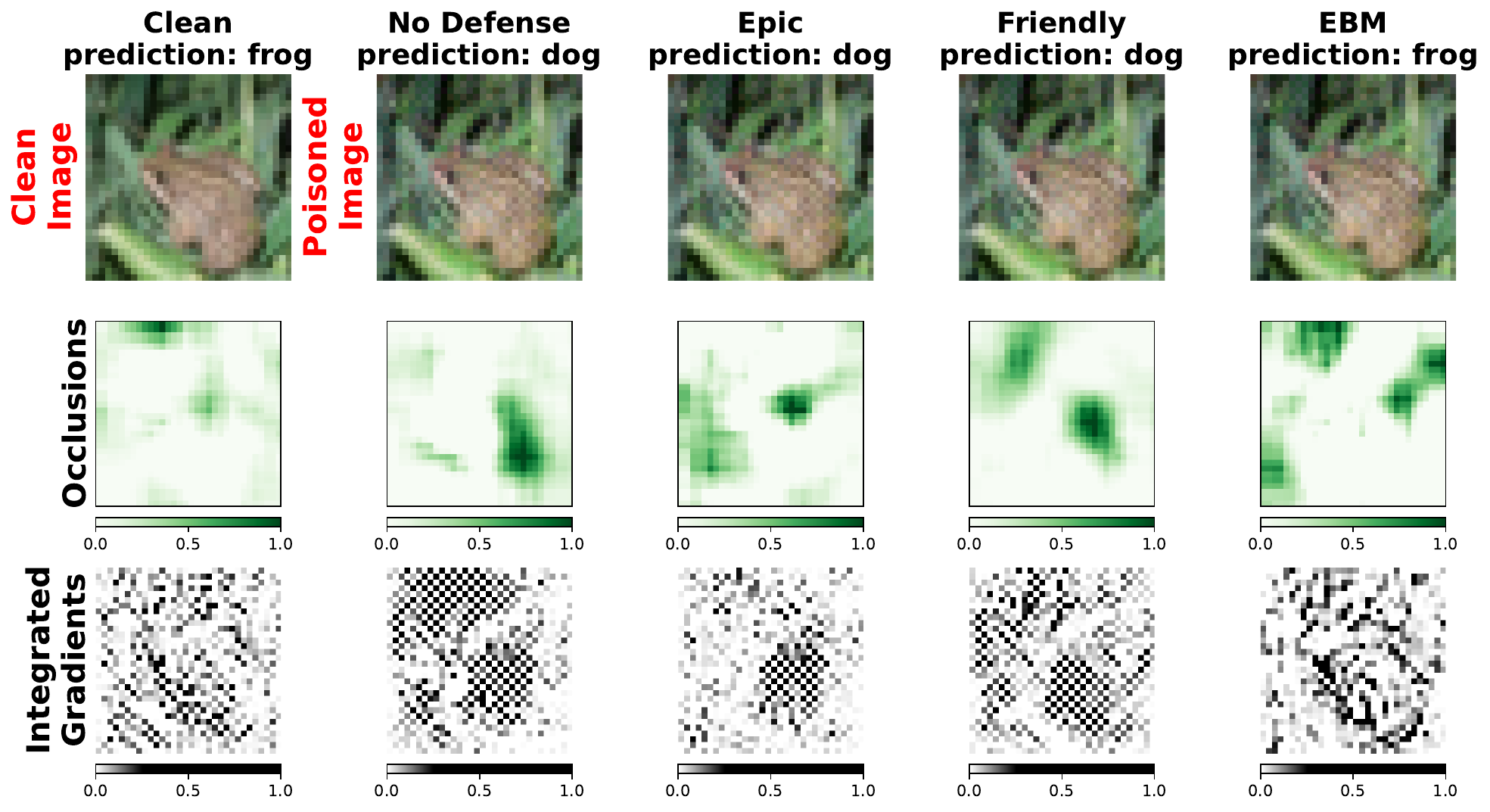}
    \end{subfigure}
    \vspace{1em} % Space between subfigures

    \begin{subfigure}{0.7\linewidth}
        \centering
        \includegraphics[width=\linewidth]{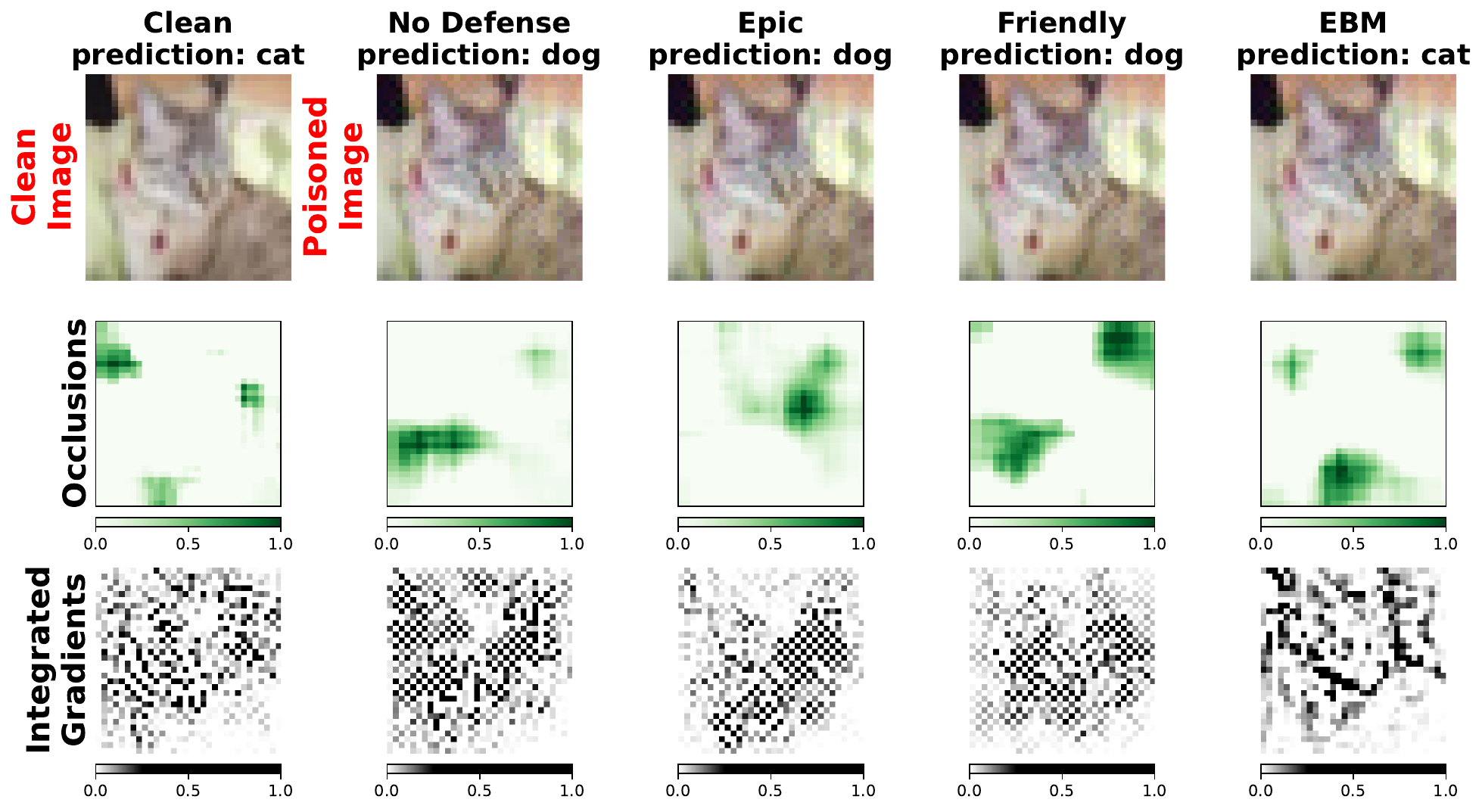}
    \end{subfigure}
    \vspace{1em}

    \begin{subfigure}{0.7\linewidth}
        \centering
        \includegraphics[width=\linewidth]{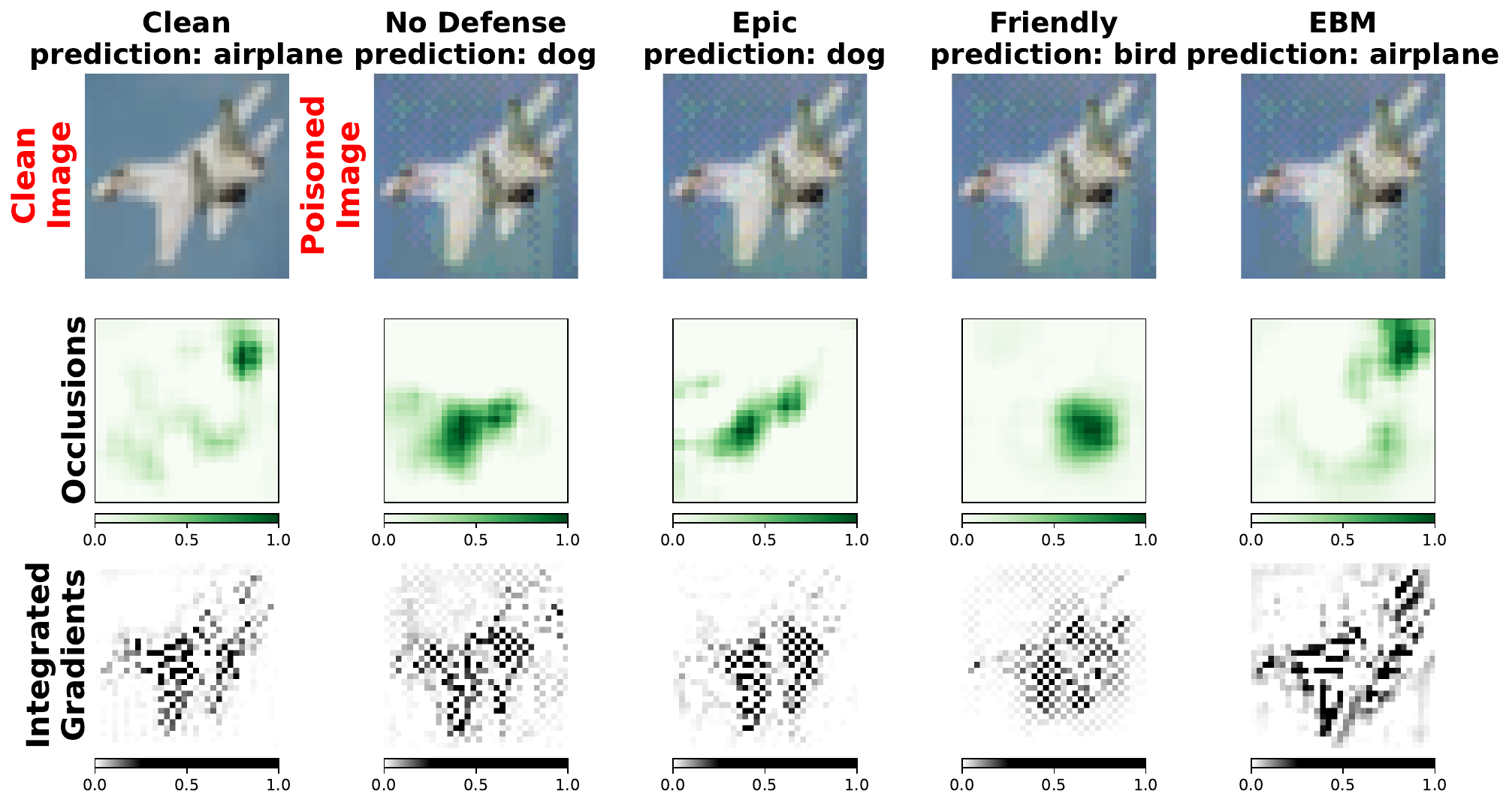}
    \end{subfigure}
    \vspace{1em}

\end{figure}

\begin{figure}[H]
    \centering
    \begin{subfigure}{0.7\linewidth}
        \centering
        \includegraphics[width=\linewidth]{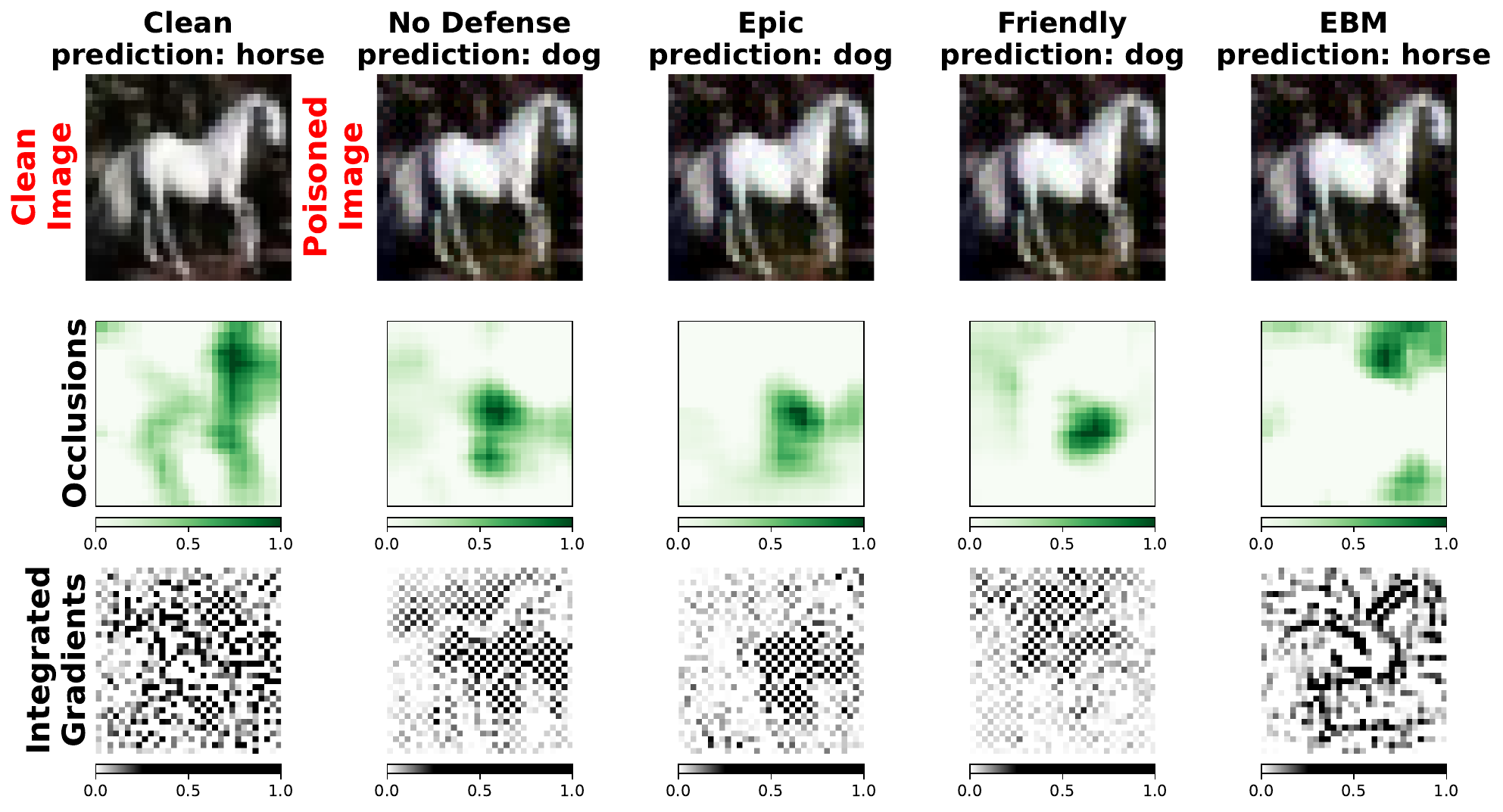}
    \end{subfigure}
    \vspace{1em}

    \begin{subfigure}{0.7\linewidth}
        \centering
        \includegraphics[width=\linewidth]{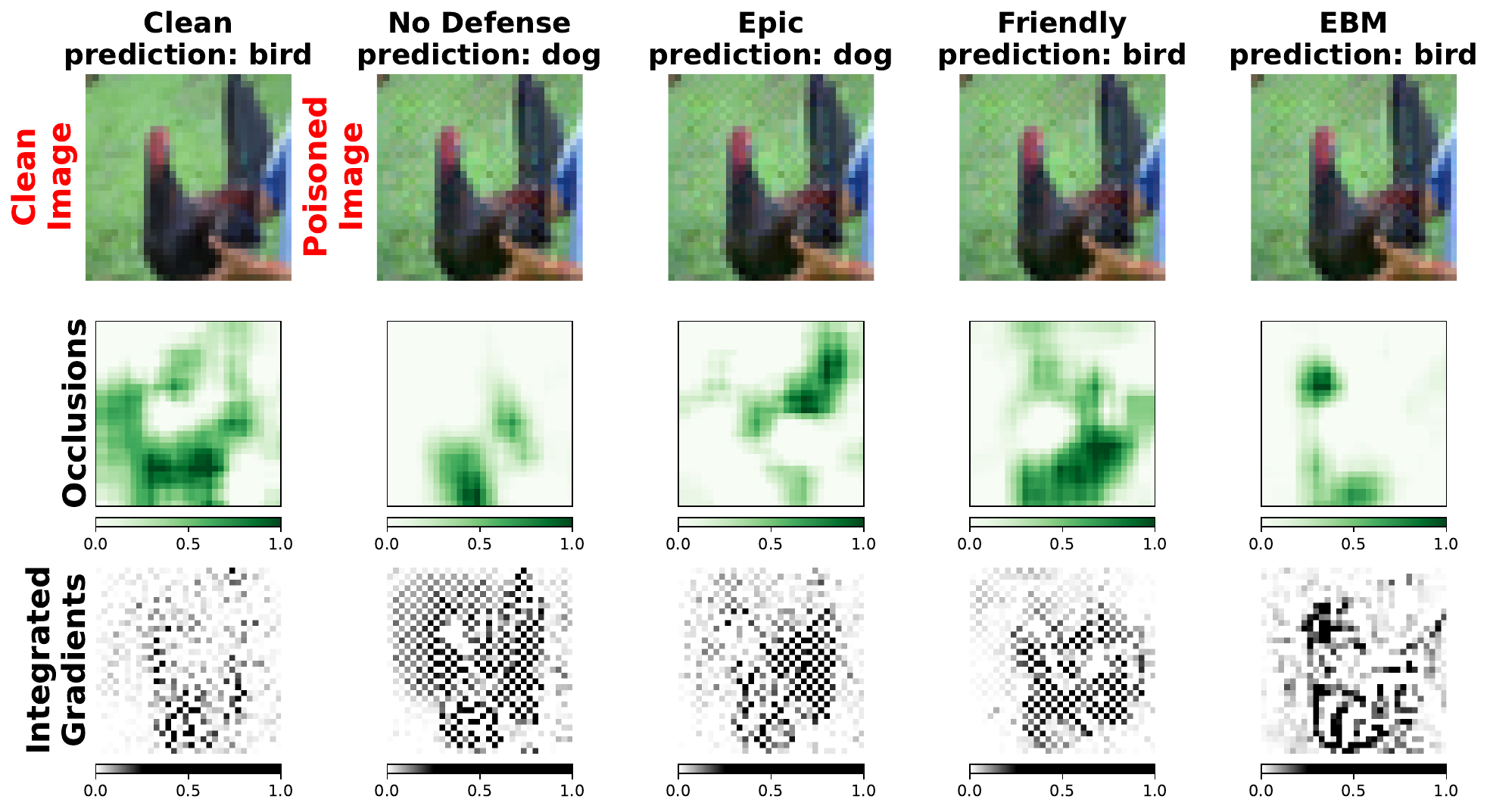}
    \end{subfigure}
    \vspace{1em}
        \begin{subfigure}{0.7\linewidth}
        \centering
        \includegraphics[width=\linewidth]{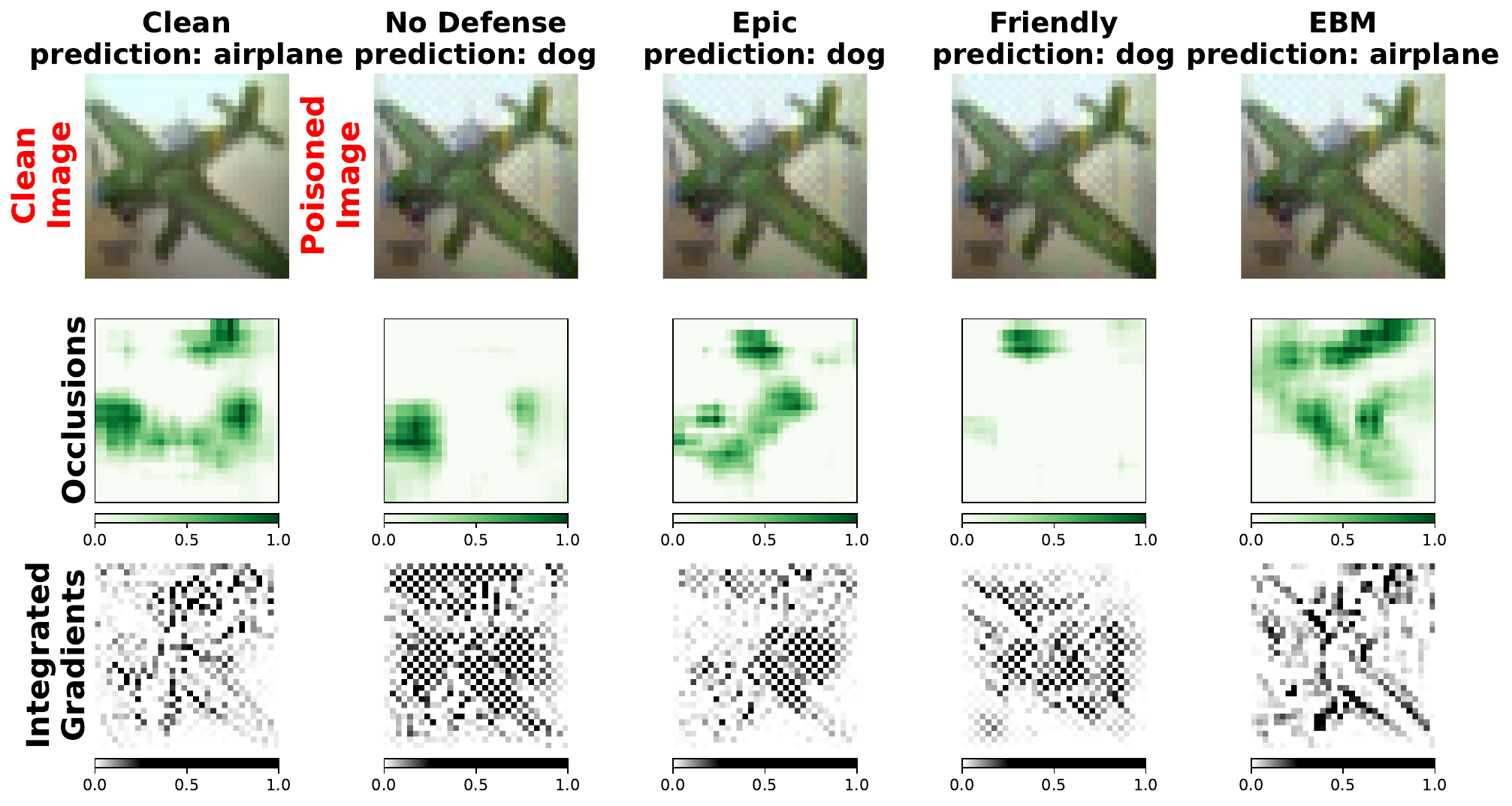}
    \end{subfigure}

\end{figure}

\newpage
\subsection{Poisoned Parameters Diverge} 
\label{app:poi-diverge}
\begin{figure}[ht]
    \centering
    \includegraphics[width=0.8\linewidth]{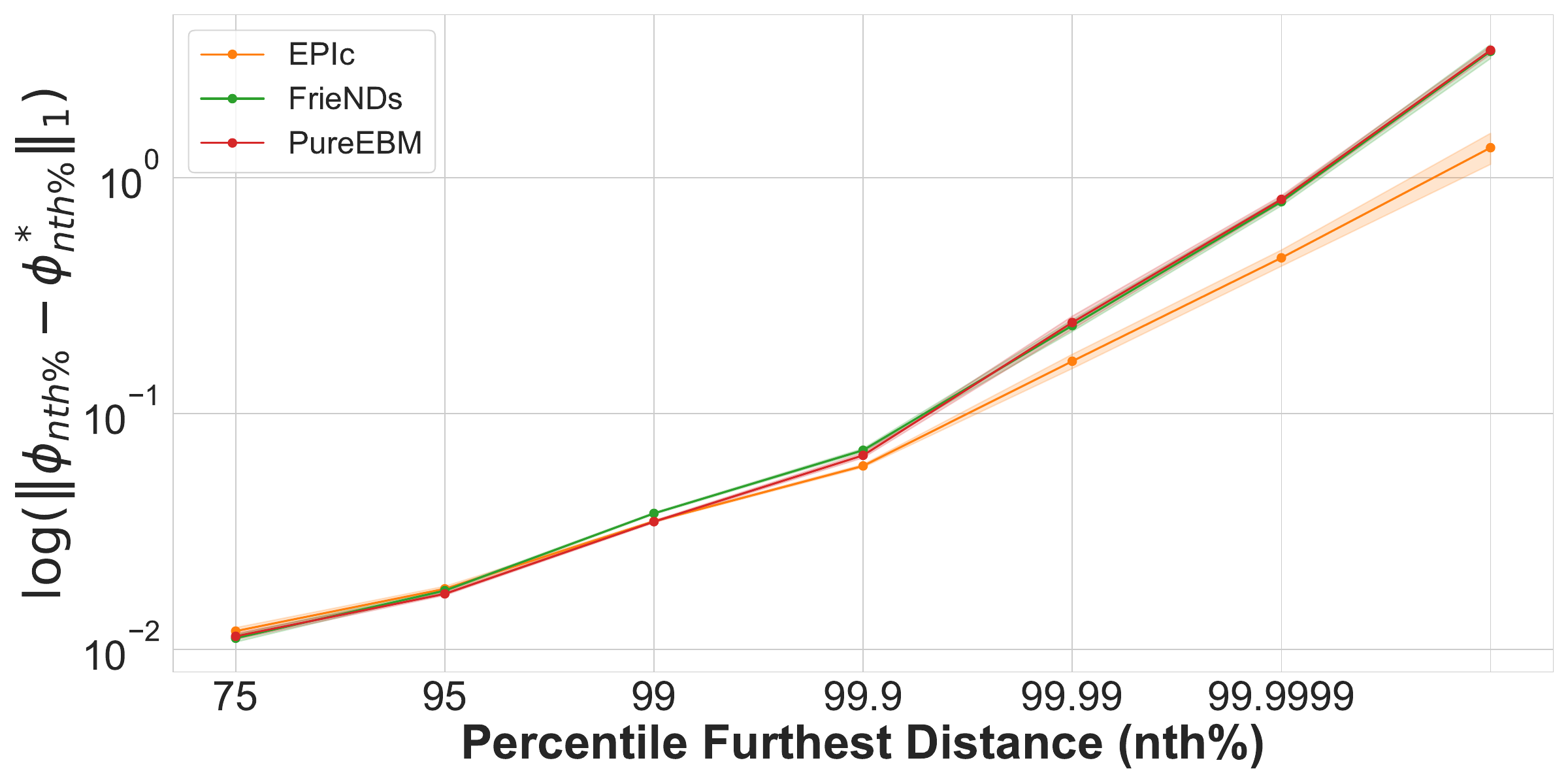}
    \caption{Comparing parameter distances from defended models to poisoned model (same init) for increasingly higher percentiles of the most moved parameters. \pebm trained models show the least movement in the tail of parameter which poisons are theorized to impact most (followed very closely by \friends but well above \epic).}
    \label{fig:param_dist_pct}
\end{figure}

\cite{yang2022poisons} proposes a subset selection method \epic which rejects poison points through training. 
This defense method produces coresets, that under the PL* condition ($\frac{1}{2}\|\mathbf{\nabla_{\phi }}\mathcal{L}(\phi)\|^{2} \geq \mu\mathcal{L}(\phi), \forall \phi$), when trained on converges to a solution $\phi^*$ with similar training dynamics to that of training on the full dataset. While such a property is attractive for convergence guarantees and preserving the overall performance of the NN, converging with dynamics too close to the poisoned parameters may defeat the purpose of a defense. As such we consider the closeness of a defended network's parameters $\phi^*$ to a poisoned network's parameters $\phi$ by measuring the L1 distance at the end of training ($\left \| \phi-\phi^* \right\|_1$). All distances use the same parameter initialization and are averaged over 8 models from the first 8 classes of the Narcissus poison. In Figure \ref{fig:param_dist_pct}, we specifically consider increasingly higher percentiles of the parameters that moved the furthest away ($\phi_{nth\%},\phi_{nth\%}^*$). The intuition is that poisons impact only a few key parameters significantly that play an incommensurate role at inference time, and hence we would only need to modify a tail of impacted parameters to defend. As we move to increasingly higher percentiles, both the \pebm and \friends defense mechanisms show a greater distance away from the poisoned model weights, indicating significant movement in this long tail of impacted parameters. We find that, as theory predicts, defending with coresets methods yield parameters that are too close to the poisoned parameters $\phi$ leading to sub-optimal defense.

% Next we consider the parameter distance of each defended model from the poisoned model with no defense at the end of training  with same model initialization, averaging over 8 models from the first 8 classes of the Narcissus poison. In Figure \ref{fig:param_dist_pct}, we specifically consider increasingly higher percentiles of the parameters that moved the furthest away. The intuition is that poisons impact only a few key parameters significantly that play an incommensurate role at inference time, and hence we would only need to modify a long tail of parameters significantly to defend. As we go to increasingly higher percentiles, both the EBM and friendly defense mechanisms show a higher distance away from the poisoned models weights, indicating significant movement in the tail of parameters impacted. Note \epic is a defense that SGD provably converges to a solution $w$ that is within an $\upsilon$-ball of the $w*$ (global minimum of $\mathcal{L}$ ) when SGD is ran on its subsets. 

\newpage

\section{EBM Langevin Dynamics Grid Searches}\label{app:grid_search}

\begin{figure}[H]
    \centering
    \includegraphics[width=0.8\linewidth]{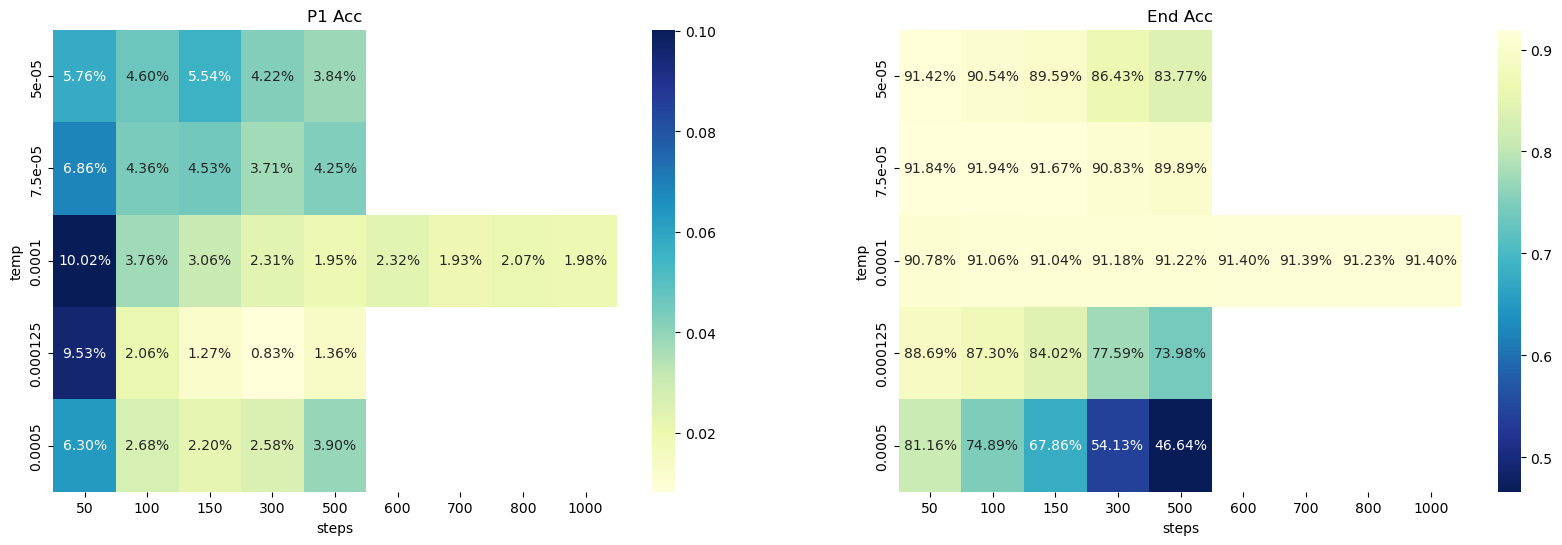}
    \caption{Grid Search for Langevin steps and temp on Narcissus Fine-Tune Transfer}
    \label{fig:grid_NS_FT}
\end{figure}

\begin{figure}[H]
    \centering
    \includegraphics[width=0.8\linewidth]{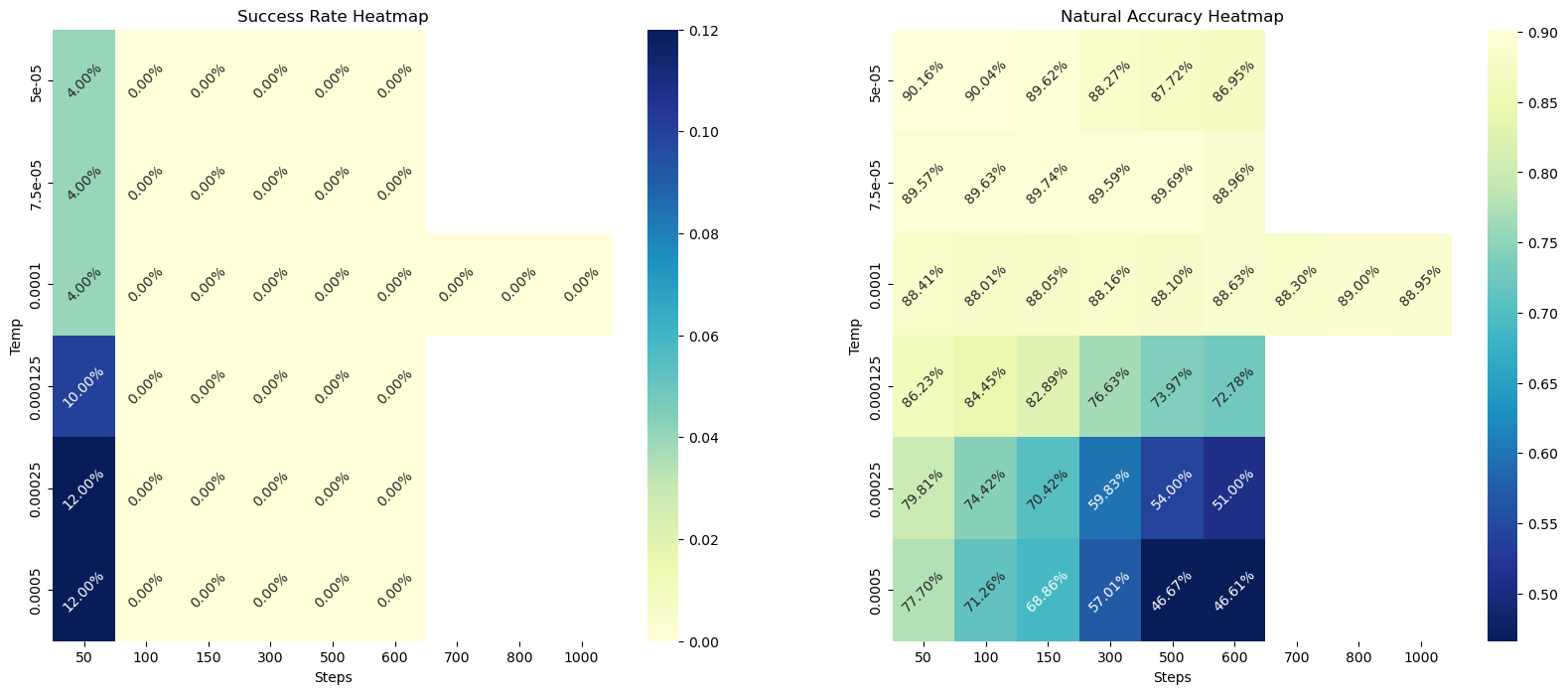}
    \caption{Grid Search for Langevin steps and temp on Bullseye Polytope Fine-Tune Transfer}
    \label{fig:grid_BP_FT}
\end{figure}

\begin{figure}[H]
    \centering
    \includegraphics[width=0.8\linewidth]{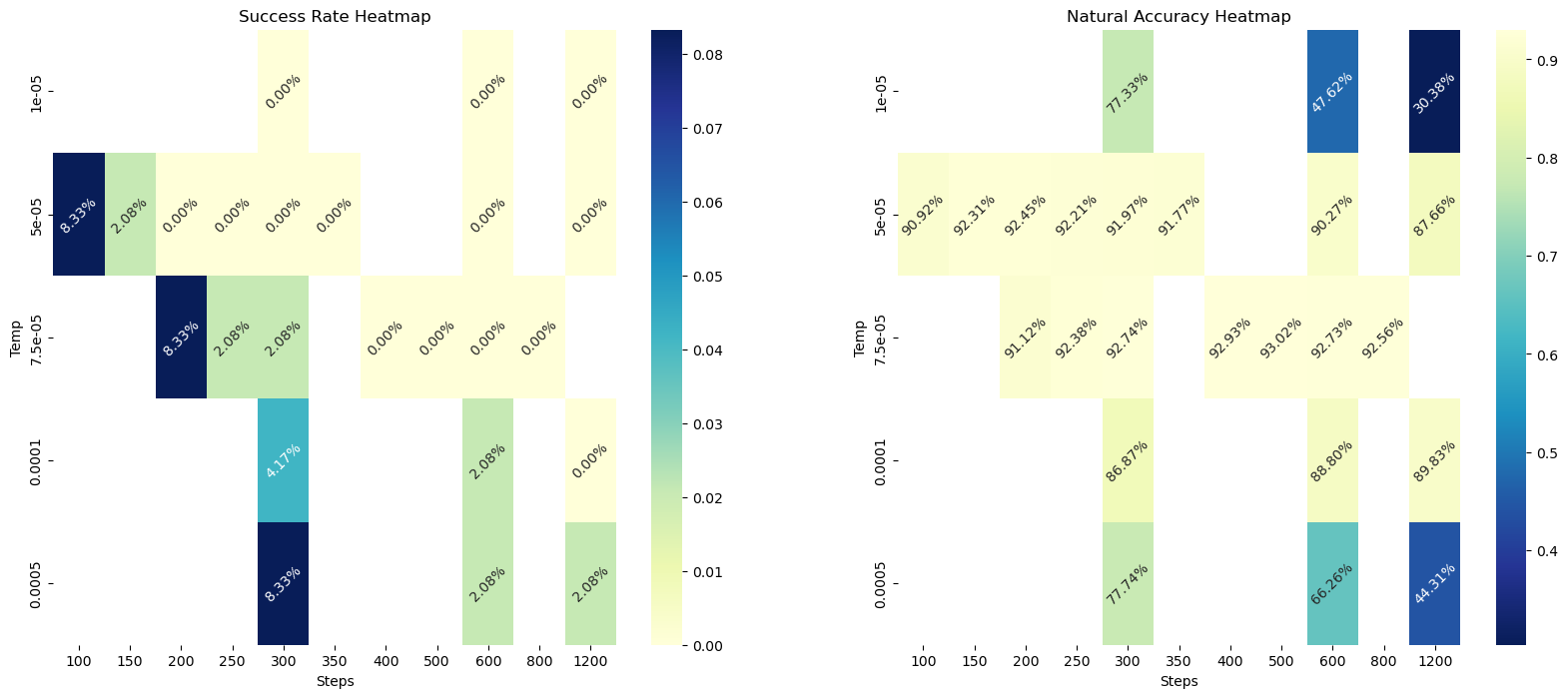}
    \caption{Grid Search for Langevin steps and temp on Bullseye Polytope Linear Transfer}
    \label{fig:grid_BP_LT}
\end{figure}

\section{Poisoned \pebm}

\begin{figure}[htbp]
    \centering
    \begin{minipage}{.48\textwidth}
        \centering
        \includegraphics[width=\linewidth]{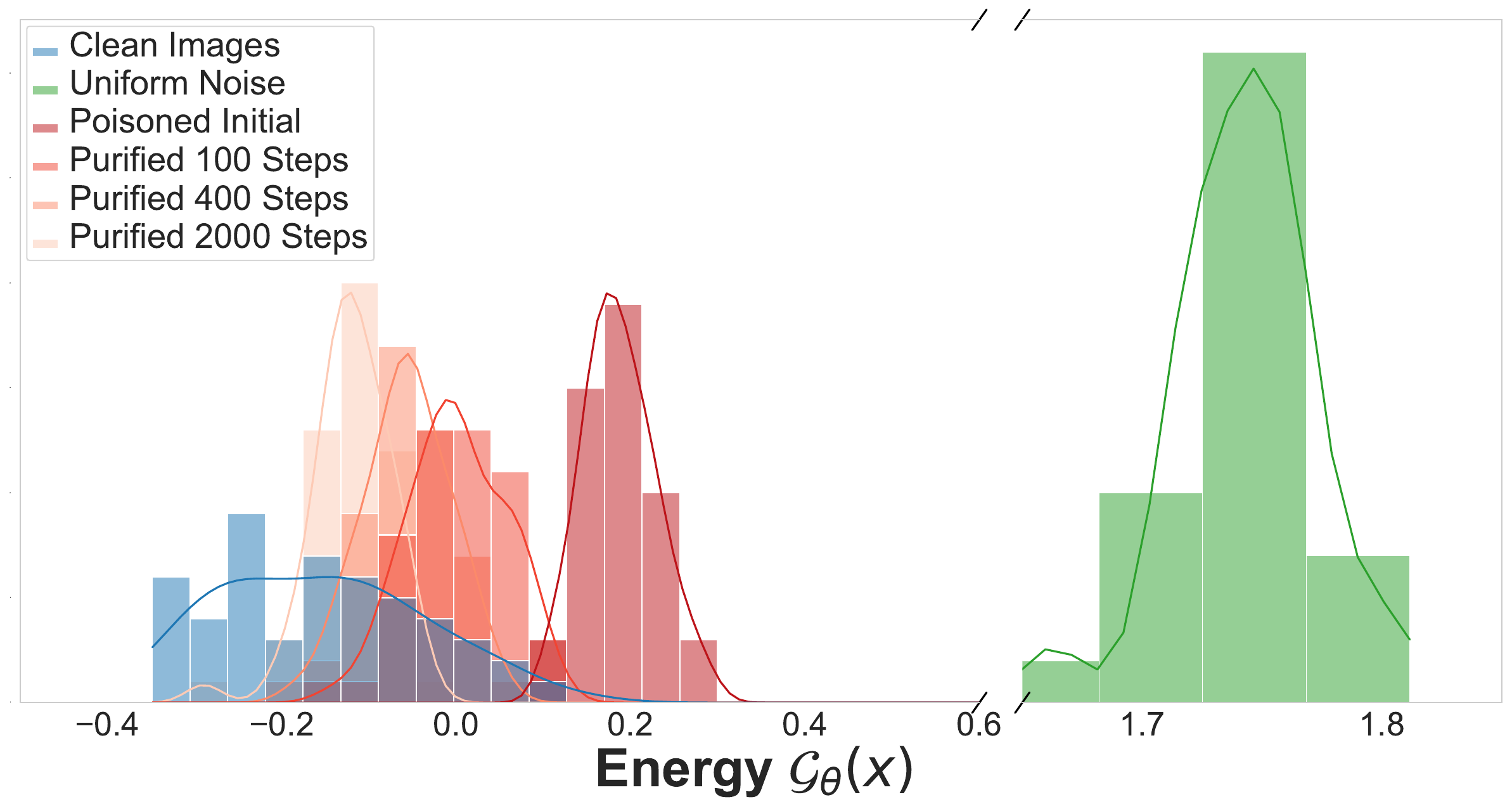}
        % \label{fig:poisonedPEBM}
    \end{minipage}\hfill
    \begin{minipage}{.48\textwidth}
        \centering
        \begin{tikzpicture}
            \node[anchor=south west,inner sep=0] (image) at (0,0) {
                \includegraphics[width=\linewidth]{EBM/Figs/intro_dists.pdf}
            };
            % Draw a fat arrow
            \draw [-latex, line width=2mm, teal!45] (4.5,3.5) -- (1.5,2);
            % Place a text box
            \node [draw, fill=white, text width=1.8cm, align=center, above right=3.9cm and 2.8cm of image.south west] (textbox) {
                \textit{PEBM}
            };
        \end{tikzpicture}
        % \caption{Visualization of distributions with annotation using \textit{PEBM}.}
        % \label{fig:appxDists}
    \end{minipage}
        \caption{Energies of poisoned points estimated by a poisoned \pebm{} are much closer to clean points than that of poisoned points estimated by a clean \pebm{}.}    
                \label{fig:poisonedPEBM}
\end{figure}

Given a dataset $x \in \mathcal{X}$ where all samples $x$ have been poisoned, we consider what happens if we train an EBM on $\mathcal{X}.$ Specifically, we consider if the fully poisoned \pebm can 1) purify given poisoned images and 2) how the energies estimated by the poisoned \pebm compare to that of a clean \pebm. We see in \ref{fig:poisonedPEBM} that the energies predicted by a poisoned \pebm (left) are significantly closer to clean images compared to estimates from a clean \pebm (right). This offers us some insight into how the poisoned \pebm method works so effectively, counter to initial intuition. When we train a \pebm on clean images we are learning some sampling trajectory towards the maximum likelihood manifold of the clean dataset i.e. when we sample from a clean \pebm via  Langevyn Dynamics we move the input image in the direction of an expected clean image. When we train on a fully poisoned dataset it becomes unclear what should happen. Theoretically, if the poison distribution is perfectly learned, one should learn a trajectory toward a poisoned distribution. That is, if one gives a clean image to the poisoned \pebm, sampling from it should move the clean image towards the poisoned distribution, and the image could become poisoned itself. Another byproduct is that poisoned images, since they have been trained on, should have a low energy. From Figure \ref{fig:poisonedPEBM} left we see that the energies of the poisoned images are much lower than that of Figure \ref{fig:intro_figure}, reproduced here (Fig. \ref{fig:poisonedPEBM} right). 
% But th

From Tables in \ref{app:results} we see that poisoned \pebm's can perform nearly as well as clean \pebm's. This means that the reduced energy gap between poisons and clean images in this setting does not hurt the purification process. Thus, the purification process remains universal.

% \begin{figure}
%     \centering
%     \includegraphics[width=0.5\linewidth]{EBM//Figs/poisonedPEBM.png}
%     \caption{Energies of poisoned points estimated by a poisoned \pebm are much closer to clean points then that of poisoned points estimated by a clean \pebm. }
%     \label{fig:poisonedPEBM}
% \end{figure}

%     \begin{tikzpicture}
%         \node[anchor=south west,inner sep=0] (image) at (0,0) {
%             \includegraphics[width=0.5\textwidth]{EBM/Figs/intro_dists.pdf}
%         };
%         % Draw a fat arrow
%         \draw [-latex, line width=2mm, teal!45] (4.5,3.5) -- (1.5,2);
%         % Place a text box
%         \node [draw, fill=white, text width=1.8cm, align=center, above right=3.9cm and 2.8cm of image.south west] (textbox) {
%             \pebm{}
%         };
%     \vspace{-15mm}
%     \end{tikzpicture}

\section{Potential Social Impacts}
\label{app:impact}

Poisoning has the potential to become one of the greatest attack vectors to AI models. As the use of foundation models grows, the community is more reliant on large and diversely sourced datasets, often lacking the means for rigorous quality control against subtle, imperceptible perturbations. In sectors like healthcare, security, finance, and autonomous vehicles, where decision making relies heavily on artificial intelligence, ensuring model integrity is crucial. Many of these applications utilize AI where erroneous outputs could have catastrophic consequences.

As a community, we hope to develop robust generalizable ML algorithms. An ideal defense method can be implemented with minimal impact to existing training infrastructure and can be widely used. We believe that this research takes an important step in that direction, enabling practitioners to purify datasets preemptively before model training with state-of-the-art results to ensure better model reliability. The downstream social impacts of this could be profound, dramatically decreasing the impacts of the poison attack vector and increasing broader public trust in the security and reliability of the AI model. 

The poison and defense research space is certainly prone to `arms-race type' behavior, where increasingly powerful poisons are developed as a result of better defenses. Our approach is novel and universal enough from previous methods that we believe it poses a much harder challenge to additional poison crafting improvements. We acknowledge that this is always a potential negative impact of further research in the poison defense space. Furthermore, poison signals are sometimes posed as a way for individuals to secure themselves against unwanted or even malicious use of their information by bad actors training AI models. Our objective is to ensure better model security where risks of poison attacks have significant consequences. But we also acknowledge that poison attacks are their own form of security against models and have ethical use cases as well. 

This goal of secure model training is challenging enough without malicious data poisoners creating undetectable backdoors in our models. Security is central to being able to trust our models. Because our universal method neutralizes all SoTA data poisoning attacks, we believe our method will have a significant positive social impact to be able to inspire trust in widespread machine learning adoption for increasingly consequential applications.

% \begin{table}[ht]
%    \centering
%    \footnotesize
%    \caption{MobileNetV2 and DenseNet121 architecture results for multiple poison types and training paradigms and HyperlightBench for a novel training paradigm the \pebm can apply to effectively.}
%    \vspace{-2mm}
%    \label{table:model_results}
%    \scalebox{0.68}{\input{EBM/Tables/models_and_datasets}}
% \end{table}

%%%%%%%%%%%%%%%%%%%%%%%%%%%%%%%%%%%%%%%%%%%%%%%%%%%%%%%%%%%%%%%%%%%%%%%%%%%%%%%
%%%%%%%%%%%%%%%%%%%%%%%%%%%%%%%%%%%%%%%%%%%%%%%%%%%%%%%%%%%%%%%%%%%%%%%%%%%%%%%

\end{document}